\documentclass{article}

\usepackage{arxiv}

\usepackage[utf8]{inputenc}
\usepackage[T1]{fontenc}

\usepackage{amsmath}
\usepackage{amssymb}
\usepackage{amsfonts}
\usepackage{amsthm}
\usepackage{bm}

\usepackage{graphicx}
\usepackage{subcaption}

\usepackage{booktabs}
\usepackage{multirow}
\usepackage{makecell}
\usepackage{tabularx}
\usepackage{array}
\usepackage{colortbl}

\usepackage{xcolor}
\usepackage{soul}

\usepackage{url}
\usepackage{hyperref}
\usepackage{cite}

\usepackage{enumitem}
\usepackage{comment}
\usepackage{latexsym}
\usepackage{nicefrac}
\usepackage{microtype}
\usepackage{lipsum}

\usepackage{algorithm}
\usepackage{algorithmic}

\usepackage{xspace}

\theoremstyle{plain}

\theoremstyle{definition}

\definecolor{routegreen}{HTML}{26734D}
\definecolor{carrierblue}{HTML}{2E6F9E}
\definecolor{centergray}{HTML}{5F6670}

\title{Green-Routed Neural Operators:\\Physics Determines Where the Network Reads}

\author{
  Chenhao Si \\
  School of Data Science \\
  The Chinese University of Hong Kong, Shenzhen \\
  Shenzhen, China \\
  \texttt{chenhao.si@link.cuhk.edu.cn}
  \And
  Ming Yan\footnotemark[1] \\
  School of Data Science \\
  The Chinese University of Hong Kong, Shenzhen \\
  Shenzhen, China \\
  \texttt{yanming@cuhk.edu.cn}
}

\begin{document}
\maketitle
\footnotetext[1]{Corresponding author.}

\begin{abstract}
We identify a mismatch between the physical role of transport fields in many PDEs and their usual role in neural operators: PDEs use them to select read coordinates, whereas neural operators typically treat them only as input values. We address this mismatch with the \emph{Green-Routed Neural Operator} (\textsc{GRNO}), which uses the governing equation to determine where latent features are sampled. A parameter-free equation adapter evaluates the diagnostic relation and constructs a departure map whose values are the read coordinates. A multiscale encoder--decoder combines centered and routed reads of latent features to learn the complete finite-time update. Across five two- and three-dimensional PDE systems, \textsc{GRNO} achieves the lowest mean final relative $L^2$ error on four under 40-step autoregressive evaluation and remains competitive on Keller--Segel. Fixed-weight route interventions reveal strong dependence on direction and spatial alignment in four systems, with weak dependence in Keller--Segel. In independently trained ablations, GRNO achieves lower mean errors than variants that supply the transport field only as an input feature, substitute a learned displacement for the equation-specified route, or apply the route with a spatial misalignment, across all five systems. It also substantially outperforms directly advecting the physical state and learning the remaining update, indicating that equation-specified read coordinates provide an effective structural prior for long-horizon PDE forecasting. Code is available at \url{https://github.com/S-Chenhao/GRNO}.
\end{abstract}

\section{Introduction}
\label{sec:introduction}

Neural operators learn maps between functions and provide fast surrogates for partial differential equations (PDEs) \cite{lu2021deeponet,li2021fno,kovachki2023neuraloperator}. During autoregressive forecasting, predictions become inputs, allowing errors to accumulate across the rollout \cite{brandstetter2022messagepassing,lippe2023pderefiner,list2025unrolling}. Existing architectures model spatial interactions through kernels, multiscale convolutions, attention, and coordinate-dependent queries; we examine how known transport fields should determine their read coordinates.

In many evolution PDEs, the equation already specifies more than which field values are relevant: it specifies the coordinates from which the evolving state should be read.  Let $q$ denote the state to be advanced, with $\mu$ specifying the physical parameters, geometry, and boundary conditions.  The current state may first determine an auxiliary field $\psi_\mu[q]$ through an instantaneous diagnostic relation, after which the equation converts $\psi_\mu[q]$ into a velocity, drift, or phase-space vector $\bm v_\mu[q]$.  We call $\psi_\mu[q]$ the \emph{diagnostic field} and $\bm v_\mu[q]$ the \emph{diagnostic carrier}. The carrier determines a departure map $\bm\Xi_{\Delta t,\mu}[q]$, whose values are the read coordinates. Sampling the previous state at these coordinates gives the transport contribution $q\circ\bm\Xi_{\Delta t,\mu}[q]$. Using $\bm v_\mu[q]$ as an input channel exposes its values at fixed grid indices, whereas using it to construct the departure map changes the coordinates from which features are read.

Endpoint supervision constrains the map $q^n\mapsto q^{n+1}$ while leaving the carrier-to-coordinate computation unspecified. Models such as Flowers make sampling locations explicit but learn their displacements from forecast data \cite{muser2026flowers}. Our mismatch lies in the computational role of the known carrier: the PDE uses it to determine read coordinates, while a generic endpoint learner leaves this role implicit.

We address this mismatch with the \emph{Green-Routed Neural Operator} (\textsc{GRNO}). A parameter-free equation adapter computes the carrier and departure map, whose values are the read coordinates used to sample latent features in a multiscale encoder--decoder. Centered and routed reads of latent features jointly determine the complete finite-time update, and the diagnostic field, diagnostic carrier, and departure map are recomputed from every predicted state. Thus, the equation determines where the network reads, while the network learns how the state evolves (Figure~\ref{fig:overview}).

Across five systems, \textsc{GRNO} achieves the lowest mean final error on four under 40-step autoregressive evaluation and remains competitive on Keller--Segel. Holding trained weights fixed and varying only the departure displacement reveals strong route dependence in four systems. Independently trained controls compare GRNO with variants that supply the transport field only as an input feature, substitute a learned displacement for the equation-specified route, apply the route with a spatial misalignment, or directly advect the physical state and learn the remaining update.

This work identifies a simple mismatch---a PDE can use a known field to choose coordinates while a neural operator receives it only as values---and provides an architecture and matched experiments for correcting and testing it.

\section{Related Work}
\label{sec:related_work}
\noindent\textbf{Neural-operator interactions.}
Graph and multipole kernels model nonlocal interactions \cite{li2020graphkernel,li2020multipole}; Fourier and multiwavelet operators learn spectral interactions \cite{li2021fno,guibas2022afno,gupta2021multiwavelet}. CNO and U-NO use multiscale architectures \cite{raonic2023cno,rahman2023uno}, while transformer operators learn interactions through attention and coordinate queries \cite{cao2021galerkin,wu2024transolver,wang2025cvit,liu2025nips}.
\par\vspace{0.1em}
\noindent\textbf{Representations and coordinates.}
G-FNO, SFNO, and ReViT encode spatial symmetries \cite{helwig2023gfno,bonev2023sfno,wei2026revit}, RieszNO represents directional structure \cite{liu2026rno}, and RD-JEPA uses predictive latent pretraining for few-trajectory transfer across related PDEs \cite{si2026rdjepa}. Spatial transformers and deformable convolutions learn sampling transformations \cite{jaderberg2015stn,dai2017deformable}; Geo-FNO and GINO connect physical domains to latent grids \cite{li2023geofno,li2023gino}. Flowers and PARADIS learn displacements or latent advection \cite{muser2026flowers,pereira2026paradis}. GRNO constructs read coordinates from the diagnostic carrier.
\par\vspace{0.1em}
\noindent\textbf{Hybrid numerical models.}
Hybrid solvers learn discretizations \cite{barsinai2019discretizations,kochkov2021cfd}, corrections within numerical updates \cite{um2020solver}, or inference-time corrections with numerical simulation \cite{fan2026scasml}. GRNO uses the physical transport computation to select latent features, while learning the complete finite-time update.
\par\vspace{0.1em}
\noindent\textbf{Green structure.}
Prior work learns Green kernels, constructs Green-inspired elliptic operators, or embeds discrete Green superposition \cite{boulle2022greenlearning,boulle2022learninggreens,melchers2025ngo,wang2024beno,tang2026gsno,tan2026dgnet}. GRNO uses the known Green inverse of a diagnostic relation to construct read coordinates for latent features.

\section{Motivation: The PDE determines where information is read}
\label{sec:motivation}
\noindent\textbf{Problem setup.}
Given fully observed training trajectories of a PDE with known coefficients and boundary conditions, we learn a map $\mathcal M_\theta$ from the current field $q^n=q(t_0+n\Delta t)$ to the next field. Forecasting starts from one exact frame and proceeds recursively: $\widehat q^0=q^0$ and $\widehat q^{n+1}=\mathcal M_\theta(\widehat q^n)$, with prescribed static inputs omitted from the notation. Models are trained separately for each system. We use Navier--Stokes to examine how known diagnostic carriers should enter this prediction map.

\subsection{A concrete example: diagnostic routing in Navier--Stokes}
\label{sec:motivation_ns}

Let $\omega^n=\omega(t_n)$ denote the vorticity at time $t_n$, with
$\Delta t=t_{n+1}-t_n$.  On the periodic domain $\Omega=\mathbb T^2$, forced two-dimensional Navier--Stokes can be written as
\begin{align}
    &\partial_t\omega+\bm u[\omega]\!\cdot\!\nabla\omega
      =\nu\Delta\omega-\alpha\omega+f,
      \label{eq:ns_vorticity}\\
    &\Delta\psi=\omega-\overline\omega,
    \quad \bm u[\omega]=\nabla^\perp\psi .
    \label{eq:ns_diagnostic}
\end{align}
Here $\nu$ is viscosity, $\alpha$ is linear drag, $f$ is a prescribed forcing, and $\overline\omega$ is the spatial mean. The vorticity $\omega$ measures local fluid rotation and is the state being advanced. At each time, solving the Poisson equation in Eq.~(\ref{eq:ns_diagnostic}) gives the diagnostic field $\psi$ from the current $\omega$. Its rotated gradient $\nabla^\perp\psi$ gives the transporting velocity $\bm u[\omega]$, which is the diagnostic carrier.

Solving the Poisson equation and applying $\nabla^\perp$ gives
\begin{equation}
    \bm u[\omega](\bm x)
      =\int_\Omega \bm K(\bm x,\bm y)
        \bigl(\omega(\bm y)-\overline\omega\bigr)\,
        \mathrm d\bm y,
    \qquad
    \bm K(\bm x,\bm y)=\nabla_{\bm x}^{\perp}G(\bm x,\bm y),
    \label{eq:ns_biot_savart}
\end{equation}
where $G$ is the mean-zero periodic Green function and $\bm K$ is the associated Biot--Savart kernel.  This kernel weights vorticity contributions across the domain to obtain the velocity at each target point \cite{majda2001vorticity}.

\begin{figure}[t]
\centering
\includegraphics[width=0.985\linewidth]{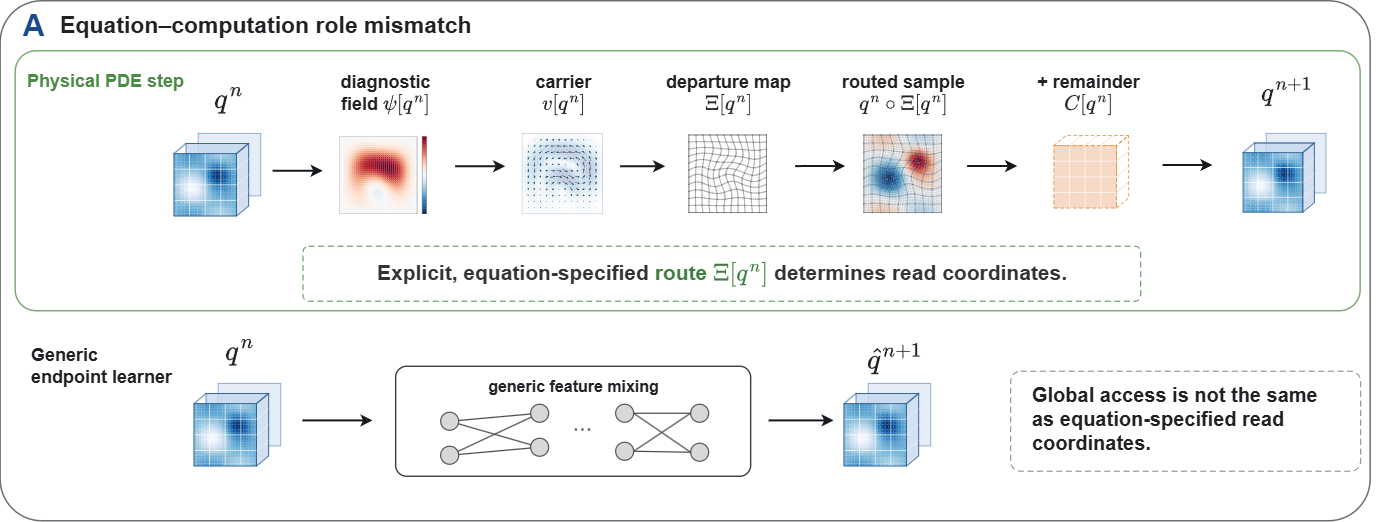}
\vspace{-2mm}
\includegraphics[width=0.985\linewidth]{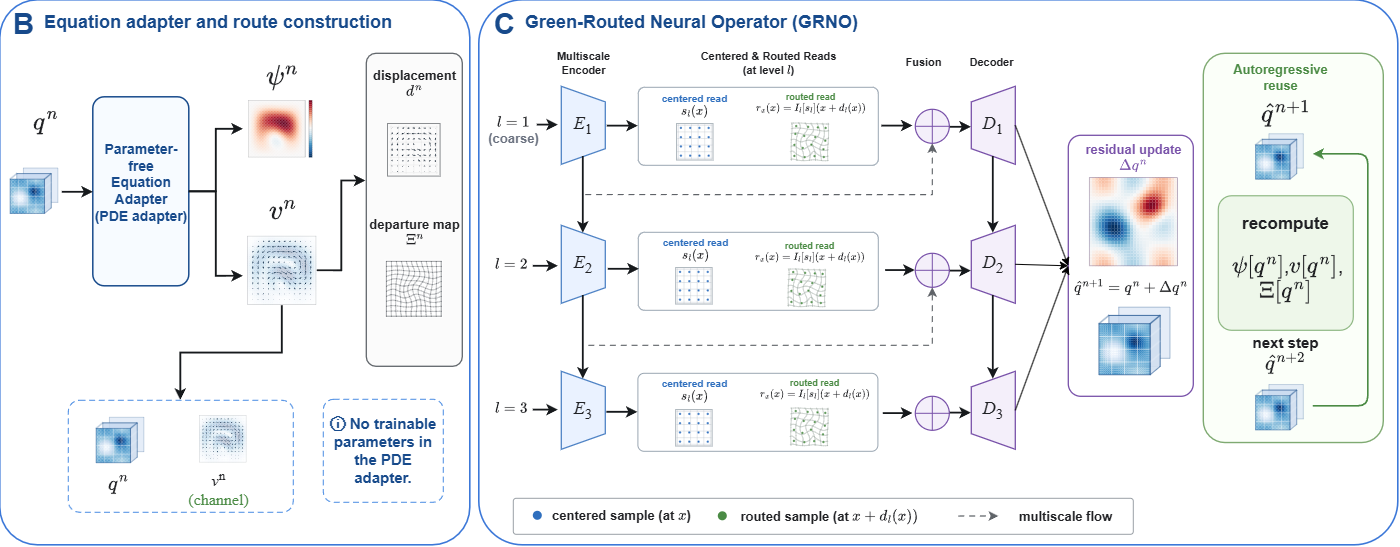}
\caption{\textbf{Equation-specified routing and GRNO.}
\textbf{(A)} The current state $q^n$ determines a diagnostic field $\psi[q^n]$ and a diagnostic carrier $\bm v[q^n]$. Backtracing the carrier gives the departure map $\bm\Xi[q^n]$, which selects the routed physical sample $q^n\!\circ\!\bm\Xi[q^n]$; $\mathcal C[q^n]$ is the remainder containing all other finite-time effects. GRNO explicitly realizes the carrier-to-coordinate-to-sample chain by using the carrier to construct read coordinates.
\textbf{(B)} The parameter-free equation adapter evaluates the known diagnostic relation, supplies the carrier as an encoder feature, and uses a backward trace to construct the displacement $\bm d^n$ and departure map $\bm\Xi^n$.
\textbf{(C)} A local multiscale encoder produces features $s_\ell$. At each scale, GRNO combines the centered read $s_\ell(\bm x)$ with the routed read
$r_\ell(\bm x)=\mathcal I_\ell[s_\ell](\bm x+\bm d_\ell(\bm x))$.
The decoder fuses the two reads, predicts the complete state increment, and recomputes the diagnostic field, diagnostic carrier, and departure map from every predicted state during autoregressive rollout.}
\label{fig:overview}
\end{figure}

Now fix a target coordinate $\bm x$ at time $t_{n+1}$.  Backtracking the velocity field defines a departure map $\bm\Xi_{\Delta t}^{n}$, where
$\bm\Xi_{\Delta t}^{n}(\bm x)$ is the earlier coordinate of the material arriving at $\bm x$.  Along this characteristic,
\begin{equation}
\begin{aligned}
    \omega^{n+1}(\bm x)
      &=\omega^n\!\left(\bm\Xi_{\Delta t}^{n}(\bm x)\right)
        +\mathcal C_{\Delta t}^{n}(\bm x),\\
    \bm\Xi_{\Delta t}^{n}(\bm x)
      &=\bm x-\Delta t\,\bm u^n(\bm x)+\mathcal O(\Delta t^2),
      \qquad
      \bm u^n:=\bm u[\omega^n].
\end{aligned}
\label{eq:ns_characteristic_identity}
\end{equation}
The pullback $\omega^n\circ\bm\Xi_{\Delta t}^{n}$ reads the old vorticity at the departure coordinate, while $\mathcal C_{\Delta t}^{n}$ collects viscosity, forcing, drag, and the remaining finite-time effects \cite{staniforth1991semi}. For intuition, pure uniform rightward transport by one grid spacing makes each target read its left neighbor. The departure map extends this lookup to spatially varying velocities. The update therefore contains the explicit sequence
\begin{equation}
    \omega^n\xrightarrow{\text{Biot--Savart}}\bm u^n\xrightarrow{\text{backtrace}}\bm\Xi_{\Delta t}^{n}\xrightarrow{\text{sample}}\omega^n\circ\bm\Xi_{\Delta t}^{n}\xrightarrow{+\,\mathcal C_{\Delta t}^{n}}\omega^{n+1}.
    \label{eq:ns_computational_chain}
\end{equation}
Appending $\bm u^n$ as an input channel exposes its values at fixed indices. Eq.~(\ref{eq:ns_characteristic_identity}) uses the same velocity differently: it moves the read coordinates before vorticity is read. Figure~\ref{fig:overview}A illustrates this value-versus-coordinate distinction.

\subsection{Where the mismatch arises: values versus read coordinates}
\label{sec:factorization_mismatch}

Changing the current state affects both the sampled values and the coordinates used to read them. For $\mathcal T(q):=q\circ\bm\Xi[q]$, let $D\mathcal T[q]e$ denote the first-order response to a state perturbation $e$. The chain rule gives
\begin{equation}
    D\mathcal T[q]e
    =
    \underbrace{e\circ\bm\Xi[q]}_{\text{state changes at a fixed map}}
    +
    \underbrace{(\nabla q\circ\bm\Xi[q])D\bm\Xi[q]e}_{\text{the departure map also changes}}.
    \label{eq:pullback_state_route_variation}
\end{equation}
The second term captures the response created when the carrier moves the read coordinates. In GRNO, this response enters through the departure map $\bm\Xi[q]$. When $D\bm\Xi[q]$ contains a Green inverse, a remote state perturbation can move a departure point and thereby change a local sample.

Endpoint supervision leaves the route underdetermined. For an endpoint map $\mathcal P$ and any alternative $\widetilde{\bm\Xi}[q]$, define
\begin{equation}
    \widetilde{\mathcal C}[q]
      :=\mathcal P(q)-q\circ\widetilde{\bm\Xi}[q]
    \quad\Longrightarrow\quad
    \mathcal P(q)
      =q\circ\widetilde{\bm\Xi}[q]+\widetilde{\mathcal C}[q].
    \label{eq:alternative_one_step_decomposition}
\end{equation}
With an unrestricted remainder, a change in the departure map can be offset by a change in the remainder, producing the same next state. Eq.~(\ref{eq:alternative_one_step_decomposition}) therefore admits multiple internal decompositions for identical training pairs. The governing equation specifies which departure map GRNO should use.

The mismatch is the absence of an explicit link from the known carrier to the read coordinates:
\begin{equation}
    q
    \xrightarrow{\text{diagnostic relation}}\bm v[q]
    \xrightarrow{\text{backtrace}}\bm\Xi[q]
    \xrightarrow{\text{sampling}}q\circ\bm\Xi[q].
    \label{eq:equation_specified_read_chain}
\end{equation}
Routing is most useful when the equation-specified read coordinates materially affect the finite-time read. When centered and routed reads are similar, their predictive contributions are correspondingly similar. Section~\ref{sec:experiments} measures this dependence across the five systems.

\subsection{General form: diagnostically routed PDEs}
\label{sec:diagnostically_routed_class}

The Navier--Stokes computation extends to systems with different states and diagnostic relations. Let $q:\Omega\rightarrow\mathbb R^c$ denote the evolving state, and let $\mu$ collect physical parameters, geometry, and boundary conditions. The current state first determines a diagnostic field through an instantaneous diagnostic relation,
\begin{equation}
    \mathcal A_\mu\psi_\mu[q]=\mathcal S_\mu(q),
    \label{eq:diagnostic_general}
\end{equation}
where $\mathcal S_\mu(q)$ forms the source from the current state and $\mathcal A_\mu^{-1}$ solves the diagnostic relation with the prescribed boundary and uniqueness conditions. The equation then converts the diagnostic field into a carrier,
\begin{equation}
    \bm v_\mu[q]
      =\mathcal V_\mu(q)
      :=\bm v_\mu^{\mathrm{loc}}(q)
        +\mathcal B_\mu\psi_\mu[q],
    \label{eq:drift_general}
\end{equation}
where $\bm v_\mu^{\mathrm{loc}}$ is an optional known local contribution and $\mathcal B_\mu$ converts the diagnostic field into a carrier. For Navier--Stokes, $\mathcal A_\mu=\Delta$, $\mathcal S_\mu(\omega)=\omega-\overline\omega$, $\mathcal B_\mu=\nabla^\perp$, and $\bm v_\mu^{\mathrm{loc}}=0$. The carrier enters the evolution equation as
\begin{equation}
    \partial_tq+\bm v_\mu[q]\!\cdot\!\nabla q
      =\mathcal R_\mu(q,\psi_\mu[q]),
    \label{eq:evolution_general}
\end{equation}
with $\mathcal R_\mu$ collecting diffusion, reaction, forcing, and other dynamics not represented by the transport term.  Conservative and phase-space forms are given in Appendix~\ref{app:pdes}.

When the diagnostic inverse has a Green representation, the state-to-carrier map can be written as
\begin{equation}
\begin{aligned}
    \bm v_\mu[q]
      &=\bm v_\mu^{\mathrm{loc}}(q)
        +\mathcal B_\mu\mathcal A_\mu^{-1}\mathcal S_\mu(q),\\
    \bm v_\mu[q](\bm x)
      &=\bm v_\mu^{\mathrm{loc}}(q)(\bm x)
        +\int_\Omega
          \bm K_\mu(\bm x,\bm y)
          \mathcal S_\mu(q)(\bm y)\,\mathrm d\bm y,
        \qquad
        \bm K_\mu=\mathcal B_{\mu,\bm x}G_\mu .
\end{aligned}
\label{eq:green_drift_general}
\end{equation}
Here $G_\mu$ is the Green function of the diagnostic relation.  The carrier then determines the short-time departure map
\begin{equation}
    \bm\Xi_{\Delta t,\mu}[q](\bm x)
      =\bm x-\Delta t\,\mathcal V_\mu(q)(\bm x)
       +\mathcal O(\Delta t^2).
    \label{eq:general_first_order_route}
\end{equation}

Let $\mathcal P_{\Delta t,\mu}$ denote the exact physical map over one step.  Relative to the equation-specified departure map, it can be decomposed as
\begin{equation}
    \boxed{
    \mathcal P_{\Delta t,\mu}(q)
      =q\circ\bm\Xi_{\Delta t,\mu}[q]
       +\mathcal C_{\Delta t,\mu}[q]},
    \qquad
    \mathcal C_{\Delta t,\mu}[q]
      :=\mathcal P_{\Delta t,\mu}(q)
        -q\circ\bm\Xi_{\Delta t,\mu}[q].
    \label{eq:physical_one_step_decomposition}
\end{equation}
The first term reads the state at the equation-specified read coordinates; the remainder contains all other finite-time effects. Eq.~(\ref{eq:physical_one_step_decomposition}) defines the remainder relative to the equation-specified route; choosing a different route induces a different remainder.

We call a PDE \emph{diagnostically routed} when an instantaneous diagnostic relation determines a carrier that, in turn, determines the read coordinates for the evolving state. Purely local reaction--diffusion equations lie outside this class, as do nonlocal fields that enter only as additive sources.

\subsection{How route perturbations accumulate over long rollouts}
\label{sec:route_sensitivity}

We next examine how a coordinate error changes the sampled state and affects subsequent predictions. Consider a controlled perturbation of Eq.~(\ref{eq:physical_one_step_decomposition}):
\begin{equation}
\begin{aligned}
    q^{n+1}
      &=q^n\circ\bm\Xi_n[q^n]+\mathcal C_n[q^n],\\
    \widetilde q^{\,n+1}
      &=\widetilde q^{\,n}\circ
        \bigl(\bm\Xi_n[\widetilde q^{\,n}]+\bm\eta_n\bigr)
        +\mathcal C_n[\widetilde q^{\,n}]+r_n .
\end{aligned}
\label{eq:controlled_route_defect}
\end{equation}
Here $\bm\eta_n$ perturbs the read coordinates, while $r_n$ perturbs the remainder. Assume locally that composition is bounded by $B_n$, the route is Lipschitz with constant $L_{\Xi,n}$, and the remainder is Lipschitz with constant $L_{\mathcal C,n}$. With $e_n=\widetilde q^{\,n}-q^n$ and $M_n=\|\nabla q^n\|_{L^\infty}$,
\begin{equation}
    \|e_{n+1}\|_2
      \le
      A_n\|e_n\|_2
      +M_n\|\bm\eta_n\|_2
      +\|r_n\|_2,
    \qquad
    A_n:=B_n+M_nL_{\Xi,n}+L_{\mathcal C,n}.
    \label{eq:route_sensitivity_coefficient}
\end{equation}
The term $M_n\|\bm\eta_n\|_2$ bounds the sampling error caused by a coordinate displacement; $M_n$ measures how sharply the state varies in space. The term $M_nL_{\Xi,n}$ captures feedback from an existing state error to the next departure map. Iterating the recursion gives
\begin{equation}
    \|e_N\|_2
    \le
    \Bigl(\prod_{k=0}^{N-1}A_k\Bigr)\|e_0\|_2
    +
    \sum_{n=0}^{N-1}
    \Bigl(\prod_{k=n+1}^{N-1}A_k\Bigr)
    \bigl(M_n\|\bm\eta_n\|_2+\|r_n\|_2\bigr).
    \label{eq:unrolled_route_error_bound}
\end{equation}
A route defect can propagate through later rollout steps because each predicted state determines the next diagnostic solve and departure map. Eq.~(\ref{eq:unrolled_route_error_bound}) separates the propagation of existing state error from newly introduced route and update errors. The assumptions, proof, and volume-preserving special case are provided in Appendix~\ref{app:theory}. These observations motivate an architecture that preserves the centered computation while exposing the equation-specified route as a separate latent read.
\section{Green-Routed Neural Operator}
\label{sec:method}

Figure~\ref{fig:overview}B--C illustrates the resulting design: the equation determines the read coordinates, while the neural core learns the complete finite-time update.

\subsection{Equation-specified routes}
\label{sec:method_adapter}

For state $q^n$, the parameter-free equation adapter evaluates
\begin{equation}
    \bm v^n=\mathcal V_\mu(q^n)
      =\bm v_\mu^{\mathrm{loc}}(q^n)
       +\mathcal B_\mu\mathcal A_\mu^{-1}\mathcal S_\mu(q^n),
    \label{eq:method_carrier}
\end{equation}
and, for spatial transport, constructs a midpoint departure displacement,
\begin{equation}
\begin{aligned}
    \bm y^{n+\frac12}&=\mathfrak b_\mu\!\left(\bm x-\tfrac{\Delta t}{2}\bm v^n(\bm x)\right),
    &\bm v^{n+\frac12}&=\mathcal I[\bm v^n](\bm y^{n+\frac12}),\\
    \bm d^n&=-\Delta t\,\bm v^{n+\frac12},
    &\bm\Xi^n&=\mathfrak b_\mu(\bm x+\bm d^n).
\end{aligned}
\label{eq:midpoint_route}
\end{equation}
Here $\mathcal I$ interpolates grid values at off-grid coordinates, and $\mathfrak b_\mu$ applies the coordinate boundary rule. The field $\bm v^{n+\frac12}$ is the current $\bm v^n$ sampled at $\bm y^{n+\frac12}$. Vlasov--Poisson instead uses a symmetric phase-space backtrace consisting of a half-step trace in position, a full-step trace in velocity, and a second half-step trace in position; the exact construction is given in Appendix~\ref{app:trace_details}. Periodic indices are wrapped, bounded indices are clamped, and outflow interpolation contributions are masked as specified there. The normalized neural input concatenates the state, carrier, static context, and normalized time step,
\begin{equation}
    x_\mu^n=\operatorname{cat}[\mathcal N_{q,\mu}(q^n),\mathcal N_{v,\mu}(\bm v^n),c_\mu,\Delta t/\Delta t_\mu].
    \label{eq:grno_input}
\end{equation}
Here $\Delta t_\mu$ is the reference frame spacing for system $\mu$, and $c_\mu$ denotes the prescribed static context.

\subsection{Multiscale mixing of centered and routed reads}
\label{sec:method_mixing}

Our four-scale encoder--decoder uses skip connections and local depthwise blocks \cite{ronneberger2015unet,liu2022convnext}. Its encoder produces features $s_\ell$. The displacement is average-pooled to each scale to sample the same feature tensor,
\begin{equation}
    r_\ell(\bm x)=\mathcal I_\ell[s_\ell]\!\left(\mathfrak b_{\mu,\ell}(\bm x+\bm d_\ell(\bm x))\right),
    \qquad \ell=0,\ldots,3.
    \label{eq:multiscale_routed_read}
\end{equation}
At each target grid point, $s_\ell(\bm x)$ provides the local feature vector and $r_\ell(\bm x)$ provides the vector interpolated from the departure location. The output grid stays fixed. At the coarsest scale, the decoder receives the two reads directly; at each finer scale, it combines them with the upsampled decoder feature from the next coarser scale. Both reads share one encoder; layer widths and block details are given in Appendix~\ref{app:architecture}.

\subsection{State increment, rollout, and training}
\label{sec:method_update}
\label{sec:method_training}

The finest decoder feature $z_0$ is mapped by the learned output head $H_\theta$ to predict the complete state increment $\widehat q^{\,n+1}-\widehat q^{\,n}$, including transport and the remaining dynamics:
\begin{equation}
    \widehat q^{\,n+1}=\widehat q^{\,n}
      +\Pi_\mu\!\left(\sigma_{\Delta q,\mu}H_\theta(z_0)\right),
    \label{eq:grno_update}
\end{equation}
where $\sigma_{\Delta q,\mu}$ is the fixed increment scale computed from training data and $\Pi_\mu$ enforces a known integral constraint when applicable. At the next step, Eqs.~(\ref{eq:method_carrier})--(\ref{eq:midpoint_route}) are recomputed from $\widehat q^{\,n+1}$. Training backpropagates through a $K=4$ rollout using
\begin{equation}
    \mathcal L(\theta)=Z_K^{-1}\sum_{k=1}^{K}\rho^{k-1}
    \bigl(\ell_{\mathrm{state}}^{(k)}+\lambda_\nabla\ell_\nabla^{(k)}
    +\lambda_{\mathrm{spec}}\ell_{\mathrm{spec}}^{(k)}\bigr),
    \label{eq:grno_training}
\end{equation}
where $Z_K=\sum_{k=1}^{K}\rho^{k-1}$. Loss definitions and hyperparameters are given in Appendix~\ref{app:protocol}. The diagnostic solve costs $\mathcal O(N\log N)$ for our spectral adapters; the local neural core and routed reads are linear in the number of grid points, and the equation adapter adds no trainable parameters.

\section{Experiments}
\label{sec:experiments}

We evaluate long-horizon predictive accuracy and isolate the contribution of equation-specified routing across five PDE systems.

\subsection{Benchmarks and protocol}
\label{sec:experimental_setup}

We evaluate Kolmogorov flow, dissipative SQG, Vlasov--Poisson, 3D Keller--Segel, and 3D porous convection. These systems span periodic and bounded domains, physical space and phase space, and five distinct diagnostic relations for computing the carrier (Appendix~\ref{app:pdes}).

The baselines cover four spatial interaction mechanisms. ReViT~\cite{wei2026revit} uses local canonical frames for rotationally equivariant attention. RieszNO~\cite{liu2026rno} combines directional Riesz features with spectral mixing. NIPS-AR~\cite{liu2025nips}, our autoregressive adaptation of NIPS, combines linear attention with channel-independent Fourier kernels. Flowers~\cite{muser2026flowers} learns multihead displacements for feature resampling, making it the closest coordinate-warp comparison to GRNO's equation-specified read coordinates.

Baseline comparisons use matched splits, normalization, training windows, loss, optimizer updates, and checkpoint selection. Models train through $K=4$ steps and are tested by free $H=40$ rollouts from prescribed exact frames, with each prediction reused as the next input. We average errors over 12 prescribed test-trajectory windows within each seed and report mean$\pm$standard deviation over seeds $\{0,1,2\}$ for the baseline comparison. Route interventions use seeds $\{0,1,2\}$, whereas independently trained ablations use seed~0. Full protocols and dimensional or autoregressive adaptations are specified in Appendices~\ref{app:protocol}--\ref{app:baselines}.

\subsection{Predictive utility}
\label{sec:external_comparison}

GRNO achieves the lowest mean final error on four of five systems (Table~\ref{tab:main_results}), reducing error relative to the strongest baseline by approximately $46\%$ on Kolmogorov flow, $28\%$ on SQG, $47\%$ on Vlasov--Poisson, and $80\%$ on porous convection. The strongest comparator is ReViT on SQG and Flowers on the other three tasks. On Keller--Segel, Flowers, RieszNO, and GRNO obtain errors of $0.32$, $0.34$, and $0.37$ in the table's units.

\begin{figure}[!htb]
\centering
\includegraphics[width=0.985\linewidth]{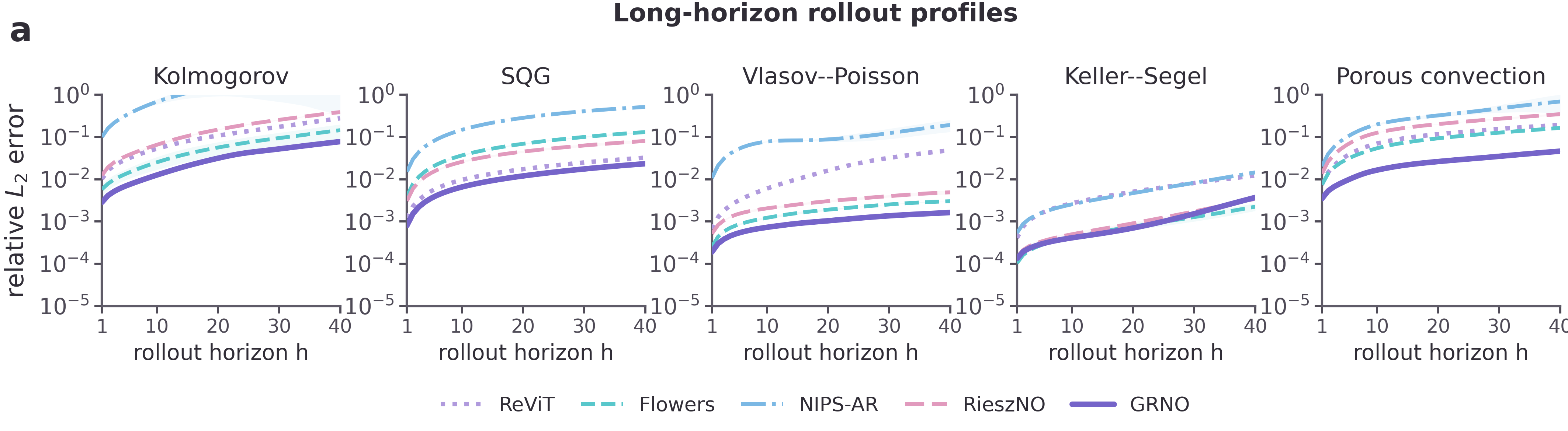}
\vspace{-2mm}
\includegraphics[width=0.985\linewidth]{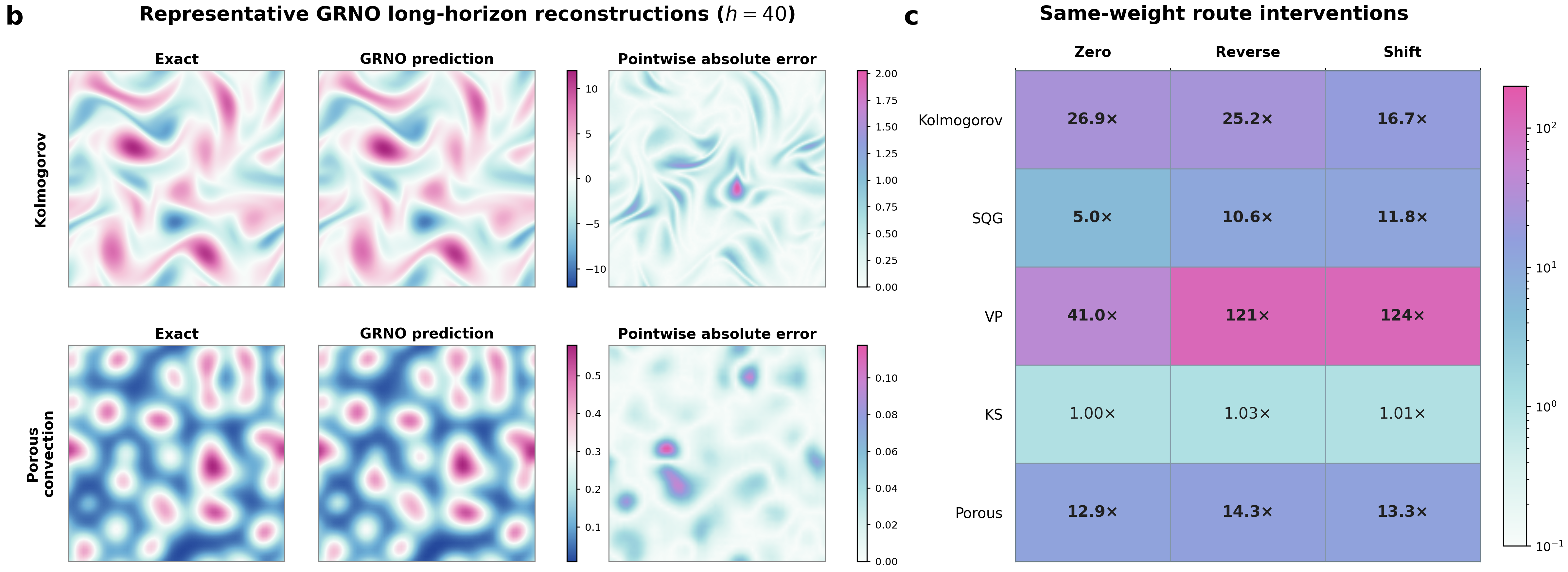}
\vspace{-2mm}
\includegraphics[width=0.985\linewidth]{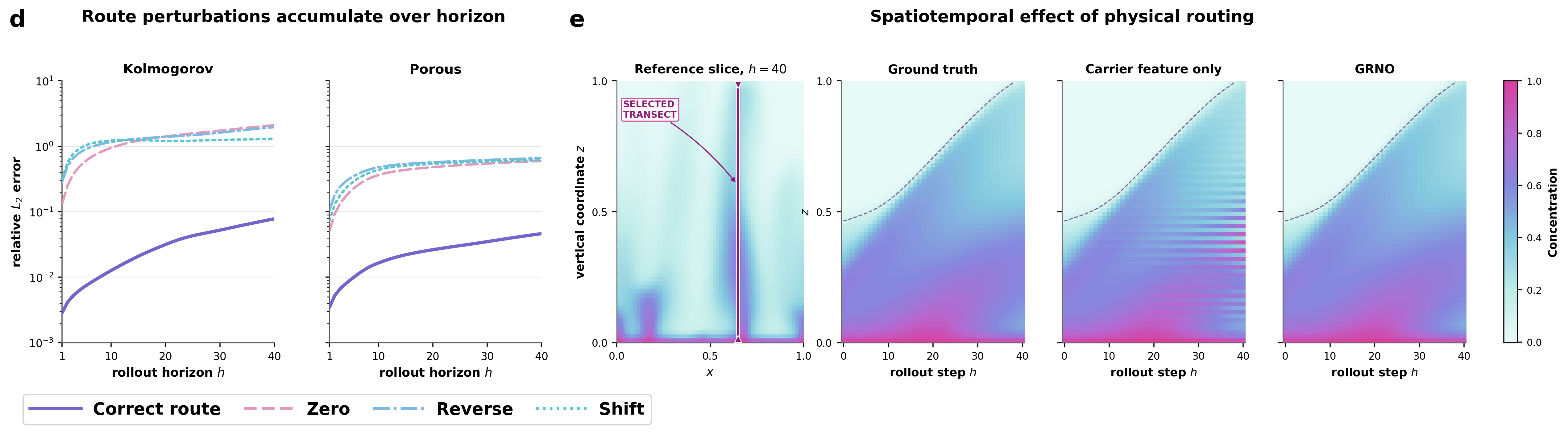}
\caption{\textbf{Long-horizon GRNO accuracy and system-dependent route use.}
\textbf{a}, Relative $L^2$ error over 40-step free rollouts. GRNO gives the lowest error on four systems and remains competitive on Keller--Segel.
\textbf{b}, Reference fields, GRNO predictions and absolute errors at $h=40$ for Kolmogorov flow and porous convection; pointwise errors remain localized around the recovered structures.
\textbf{c}, Ratios of final error under the zero, reverse, and shift interventions relative to the correct route, with network weights fixed. In four systems, intervention errors are 5.0--124 times those obtained with the correct route; in Keller--Segel, the ratios remain near one.
\textbf{d}, The error differences appear early and persist throughout the rollout.
\textbf{e}, Space--time concentration along the selected porous-convection transect for the reference, \emph{Carrier feature only}, and GRNO. In this selected window, \emph{Carrier feature only} develops temporal banding, whereas GRNO follows the advancing front.}
\label{fig:rollout_interventions}
\end{figure}

Figure~\ref{fig:rollout_interventions}a shows lower errors across substantial portions of these four rollouts, extending beyond the four supervised steps to the 40-step test horizon. On Kolmogorov flow and porous convection, GRNO also has smaller standard deviations across training seeds than Flowers: $0.56$ versus $5.39$ and $0.16$ versus $2.71$, respectively, in the units of Table~\ref{tab:main_results}.

Figure~\ref{fig:rollout_interventions}b shows that GRNO recovers the alternating vorticity regions and filaments in Kolmogorov flow and the concentration patterns in porous convection, with localized discrepancies. Appendix~\ref{app:qualitative_forecasting} compares both models' spatial errors across ten rollout horizons. The following experiments isolate the contribution of equation-specified routing.

\begin{table}[t]
\centering
\caption{\textbf{Long-horizon predictive accuracy.} Final relative $L^2$ error ($\times 10^{-2}$) at $H=40$. Entries are mean$\pm$standard deviation over three training seeds. Lower is better; bold marks the lowest mean.}
\label{tab:main_results}
\setlength{\tabcolsep}{4pt}
\resizebox{\linewidth}{!}{%
\begin{tabular}{lccccc}
\toprule
Model & Kolmogorov & SQG & Vlasov--Poisson
      & Keller--Segel & Porous convection \\
\midrule
ReViT~\cite{wei2026revit}
& $27.84\pm1.18$
& $3.25\pm0.03$
& $4.86\pm0.90$
& $1.23\pm0.02$
& $19.35\pm1.27$ \\
Flowers~\cite{muser2026flowers}
& $14.54\pm5.39$
& $13.12\pm0.50$
& $0.30\pm0.01$
& $\mathbf{0.32\pm0.09}$
& $16.51\pm2.71$ \\
NIPS-AR~\cite{liu2025nips}
& $134.25\pm 12.96$
& $51.65\pm 0.06$
& $19.33\pm 6.10$
& $1.46\pm 0.39$
& $69.33\pm 32.39$ \\
RieszNO~\cite{liu2026rno}
& $39.01\pm0.48$
& $8.07\pm0.63$
& $0.49\pm0.05$
& $0.34\pm0.01$
& $34.84\pm1.77$ \\
\textsc{GRNO} (ours)
& $\mathbf{7.80\pm0.56}$
& $\mathbf{2.34\pm0.24}$
& $\mathbf{0.16\pm0.01}$
& $0.37\pm0.02$
& $\mathbf{3.34\pm0.16}$ \\
\bottomrule
\end{tabular}%
}
\end{table}

\subsection{Does the fitted model use the route?}
\label{sec:route_interventions}

The accuracy gains above raise a direct mechanistic question: do the fitted GRNOs actually exploit the equation-specified read coordinates, or could comparable results be achieved while ignoring the route? We answer this by keeping each checkpoint and all other operations unchanged, and modifying only the departure displacement in Eq.~(\ref{eq:midpoint_route}): \emph{zero} sets $\bm d^n=0$, \emph{reverse} uses $-\bm d^n$, and \emph{shift} spatially misaligns the displacement with the current state.

\begin{table}[t]
\centering
\small
\caption{\textbf{Fixed-weight route interventions.} Final relative $L^2$ error ($\times 10^{-2}$) at $H=40$. Entries are mean$\pm$sample standard deviation across the three seed means.}
\label{tab:route_interventions}
\setlength{\tabcolsep}{6pt}
\begin{tabular}{lrrrr}
\toprule
System & Correct & Zero & Reverse & Shift \\
\midrule
Kolmogorov
& $\mathbf{7.80\pm0.56}$ & $210.18\pm 0.17$ & $196.34\pm 0.21$ & $130.67\pm 0.63$ \\
SQG
& $\mathbf{2.34\pm0.24}$ & $11.77\pm 0.36$ & $24.72\pm 0.27$ & $27.58\pm 0.31$ \\
Vlasov--Poisson
& $\mathbf{0.16\pm0.01}$ & $6.71\pm 0.09$ & $19.75\pm 0.21$ & $20.31\pm 0.06$ \\
Keller--Segel
& $\mathbf{0.37\pm0.02}$ & $0.38\pm 0.02$ & $0.39\pm 0.01$ & $0.38\pm0.01$ \\
Porous convection
& $\mathbf{3.34\pm0.16}$ & $41.08\pm 0.91$ & $47.76\pm1.17$ & $44.42\pm 1.21$ \\
\bottomrule
\end{tabular}
\end{table}

Each intervention increases mean final error by at least $16.7\times$ on Kolmogorov flow, $5.0\times$ on SQG, $41.0\times$ on Vlasov--Poisson, and $13.0\times$ on porous convection, while Keller--Segel changes little (Table~\ref{tab:route_interventions}). Figure~\ref{fig:rollout_interventions}c--d shows degradation factors and rollout profiles; Appendix~\ref{app:qualitative_route_interventions} shows the corresponding fields and pointwise errors. The fitted predictors depend strongly on route direction and alignment in four systems, and weakly in Keller--Segel.

\subsection{Does equation-specified routing improve prediction?}
\label{sec:factorial_route}

Route reliance in a fixed checkpoint does not by itself establish that equation-specified routing improves the learned predictor. We therefore retrain all variants independently to test whether equation-specified read coordinates reduce error. A $2\times2$ ablation independently enables or disables carrier features and equation-specified latent routing in the shared backbone. \emph{Centered only} uses neither component, \emph{Carrier feature only} and \emph{Route only} use one component each, and GRNO uses both. This separates the value of supplying the carrier from the value of using it to select read coordinates.

\begin{table}[t]
\centering
\small
\caption{\textbf{Independently trained ablations and route controls.} Mean final relative $L^2$ error ($\times 10^{-2}$) at $H=40$, for seed~0 and 12 prescribed test windows per system. The first four rows form the $2\times2$ factorial. Bold marks the lowest mean among evaluated variants.}
\label{tab:factorial_five_pde_seed0}
\setlength{\tabcolsep}{4pt}
\resizebox{\linewidth}{!}{%
\begin{tabular}{lrrrrr}
\toprule
Variant & Kolmogorov & SQG & Vlasov--Poisson
        & Keller--Segel & Porous convection \\
\midrule
Centered only
& 8.1659 & 4.2917 & 0.3158 & \textbf{0.3563} & 6.5352 \\
Carrier feature only
& 9.3106 & 3.7392 & 0.2849 & 0.3788 & 6.2905 \\
Route only
& 9.0460 & 3.3361 & 0.2256 & 0.3710 & 4.8331 \\
Full (GRNO)
& \textbf{7.8023} & \textbf{2.3354} & \textbf{0.1635}
& 0.3718 & \textbf{3.3426} \\
\midrule
Learned route
& 8.7553 & 2.8387 & 0.1782 & 0.3892 & 3.4775 \\
Shifted equation-specified route
& 9.5599 & 3.3667 & 0.1913 & 0.3791 & 4.0076 \\
\midrule
State advection + residual
& 11.3067 & 6.2341 & 0.6775 & 0.5907 & 9.3214 \\
\bottomrule
\end{tabular}%
}
\end{table}

With carrier features present, adding equation-specified routing reduces mean final error by $16.2\%$, $37.5\%$, $42.6\%$, $1.8\%$, and $46.9\%$ on Kolmogorov flow, SQG, Vlasov--Poisson, Keller--Segel, and porous convection, respectively (Table~\ref{tab:factorial_five_pde_seed0}). GRNO achieves the lowest mean on four systems, whereas \emph{Centered only} leads on Keller--Segel. On SQG and Vlasov--Poisson, either \emph{Carrier feature only} or \emph{Route only} improves on \emph{Centered only}, and GRNO improves further. On Kolmogorov flow, the improvement over \emph{Centered only} emerges when carrier features and routing are combined. Figure~\ref{fig:rollout_interventions}e compares \emph{Carrier feature only} with GRNO along the same porous-convection transect.

We next test how the read coordinates are constructed. The \emph{Learned route} control replaces the equation-specified departure displacement with a displacement head trained using the forecasting loss. The \emph{Shifted equation-specified route} control spatially misaligns the equation-specified departure displacement with the current state. Both retain carrier features and are independently trained using their respective read coordinates. GRNO has lower mean errors than both controls on all five systems.

Finally, we compare routing the physical state with routing latent features. The \emph{State advection + residual} control uses the same equation-specified read coordinates to resample the physical state and then applies a centered network to learn the remaining update, without latent routing. GRNO reduces final error relative to this control by $31.0\%$, $62.5\%$, $75.9\%$, $37.1\%$, and $64.1\%$ on Kolmogorov flow, SQG, Vlasov--Poisson, Keller--Segel, and porous convection, respectively. \emph{State advection + residual} has the highest mean error among the evaluated variants on all five systems. These ablations indicate that the gain depends on using the equation-specified route for latent feature reads while learning the complete finite-time update.

\section{Discussion}
\label{sec:discussion}
GRNO uses the diagnostic carrier both as an input feature and to determine where latent features are read. The diagnostic relation determines the diagnostic field, from which the carrier and departure map are computed. The values of the departure map are the read coordinates used by the neural operator. The equation therefore specifies the read coordinates, while the network learns the complete finite-time update. This interface applies across two-dimensional flow, phase-space dynamics, and three-dimensional transport.

Fixed-weight route interventions show that four fitted predictors depend strongly on route direction and alignment. In the independently trained controls, GRNO has lower mean error than \emph{Learned route} and \emph{Shifted equation-specified route} on all five systems. It also has lower mean error than \emph{State advection + residual} on every benchmark. These results indicate that performance depends on both supplying the carrier and using it to determine latent read coordinates.

Dependence on routing varies by system. Keller--Segel shows little sensitivity to fixed-weight route interventions. In the ablations, GRNO improves only slightly on \emph{Carrier feature only}, and \emph{Centered only} has the lowest error. In the other four systems, the trained models are sensitive to route interventions, and GRNO has the lowest mean error in the ablations. These results support equation-specified latent routing when the route materially affects finite-time prediction.

\subsection*{AI use statement}
The authors used generative AI tools to help polish the text, check the mathematical arguments, give feedback on the experimental design and controls, implement and debug the model code, and write the plotting code based on the model code from their trained models and the requirements they gave. 

The authors conducted all numerical experiments and reviewed the AI-assisted code, text, and figures. All reported numerical results were obtained from these experiments. The authors take responsibility for the final content of this work.

\subsection*{Ethics statement}

This work uses synthetic numerical PDE data and does not involve human subjects, personal data, or sensitive datasets.

\section*{Reproducibility Statement}
All equations, datasets, adapters, boundary rules, model configurations, losses, training budgets, and evaluation metrics are specified in the appendices. The code will be shared after the paper is accepted.

\bibliographystyle{unsrt}  
\bibliography{ref}

\clearpage
\appendix
\section{Proofs for the route-sensitivity bounds}
\label{app:theory}
Assume that, in a neighborhood of the reference trajectory,
\begin{align}
  \|f\circ(\bm\Xi_n[p]+\bm\eta)\|_2&\le B_n\|f\|_2,\\
  \|\bm\Xi_n[p]-\bm\Xi_n[q]\|_2&\le L_{\Xi,n}\|p-q\|_2,\\
  \|\mathcal C_n[p]-\mathcal C_n[q]\|_2&\le L_{\mathcal C,n}\|p-q\|_2.
\end{align}
Let $e_n=\widetilde q^{\,n}-q^n$ and $M_n$ bound $\|\nabla q^n\|_{L^\infty}$ along the segment joining the two departure maps.  Add and subtract $q^n\circ(\bm\Xi_n[\widetilde q^{\,n}]+\bm\eta_n)$ in Eq.~(\ref{eq:controlled_route_defect}).  The first resulting term is bounded by $B_n\|e_n\|_2$; the mean-value theorem and the second inequality bound the coordinate change by $M_n(L_{\Xi,n}\|e_n\|_2+\|\bm\eta_n\|_2)$; the final inequality bounds the remainder difference.  Summing the three bounds gives the one-step recursion in Eq.~(\ref{eq:route_sensitivity_coefficient}).  Repeated substitution proves Eq.~(\ref{eq:unrolled_route_error_bound}).  For a measure-preserving continuum characteristic, change of variables gives $B_n=1$; bounded-domain handling and discrete interpolation are covered by retaining $B_n$.

\section{Supplementary guide}
\label{app:guide}

The supplement is organized around the claims made in the main paper. Appendix~\ref{app:theory} proves the route-sensitivity bounds.  Appendix~\ref{app:pdes} instantiates the diagnostically routed PDE class and specifies every dataset. Appendix~\ref{app:trace_details} gives the parameter-free equation adapter, traces, and boundary rules. Appendix~\ref{app:architecture} records the shared trainable architecture, and Appendix~\ref{app:protocol} fixes the training and evaluation contract. Appendix~\ref{app:baselines} documents baseline provenance and adaptations. Appendices~\ref{app:metrics}--\ref{app:route_study} define the reported metrics and describe both the fixed-weight interventions and independently trained controls. Appendix~\ref{app:additional_visualization} provides forecasting and route-intervention visualizations.
\begin{table}[h]
    \centering
    \caption{\textbf{Claim--evidence map.}  This table separates evidence for
    predictive utility from evidence for the proposed routing mechanism.}
    \label{tab:claim_evidence}
    \small
    \setlength{\tabcolsep}{5pt}
    \begin{tabularx}{\textwidth}{>{\raggedright\arraybackslash}p{0.23\textwidth}
                                  >{\raggedright\arraybackslash}X
                                  >{\raggedright\arraybackslash}p{0.25\textwidth}}
        \toprule
        Question & Control & Evidence \tabularnewline
        \midrule
        Does the complete model predict well?
        & Independently trained models share data, loss, update budget,
          checkpoint rule, and evaluator.
        & Main Table~\ref{tab:main_results} \tabularnewline
        Does the trained model use the route?
        & Keep weights and carrier features fixed; change only the departure displacement.
        & Figure~\ref{fig:rollout_interventions} and Appendix~\ref{app:route_study} \tabularnewline
        Does equation-specified routing improve prediction after training?
        & Independently train models with different carrier features,
          coordinate constructions, and routing targets.
        & Main Table~\ref{tab:factorial_five_pde_seed0} and
          Appendix~\ref{app:route_study} \tabularnewline
        \bottomrule
    \end{tabularx}
\end{table}

\section{PDE instantiations and datasets}
\label{app:pdes}

\subsection{Variants of the common template}

Eq.~(\ref{eq:diagnostic_general})--(\ref{eq:evolution_general}) use an advective notation for readability.  When the evolved quantity is a density, the corresponding conservative form is
\begin{equation}
    \partial_t q+\nabla\!\cdot\!\bigl(\bm v_\mu[q]q\bigr)
    =\mathcal R_\mu(q,\psi_\mu[q]).
    \label{eq:conservative_template}
\end{equation}
For a phase-space state $f(\bm x,\bm v,t)$, the same idea applies on the product domain:
\begin{equation}
    \partial_t f+\bm a_x[f]\!\cdot\!\nabla_x f
    +\bm a_v[f]\!\cdot\!\nabla_v f=\mathcal R(f),
    \label{eq:phase_space_template}
\end{equation}
where the diagnostic solve may determine either component of the phase-space carrier $(\bm a_x,\bm a_v)$; Table~\ref{tab:carrier_instances} gives the exact instantiation for every task.

\begin{table}[t]
    \centering
    \caption{\textbf{Diagnostic carriers used in the experiments.} Each row lists the evolved state, diagnostic relation, diagnostic carrier that determines the read coordinates, and boundary rule.}
    \label{tab:carrier_instances}
    \scriptsize
    \setlength{\tabcolsep}{3.5pt}
    \resizebox{\textwidth}{!}{%
    \begin{tabular}{lllll}
        \toprule
        System & Evolved state $q$ & Diagnostic relation & Diagnostic carrier
               & Boundary rule \\
        \midrule
        Kolmogorov
        & $\omega$
        & $\Delta\psi=\omega-\bar\omega$
        & $\bm u=\nabla^\perp\psi$
        & periodic $x,y$ \\
        SQG
        & $\theta$
        & $\psi=-(-\Delta)^{-1/2}\theta$
        & $\bm u=\nabla^\perp\psi$
        & periodic $x,y$ \\
        Vlasov--Poisson
        & $f(x,v)$
        & $\rho=\int f\,\mathrm dv$, $\partial_xE=1-\rho$
        & $(a_x,a_v)=(v,-E)$
        & periodic $x$; zero outflow in $v$ \\
        Keller--Segel
        & $\rho$
        & $(-\Delta+\lambda^2)c=\rho-\bar\rho$
        & $\bm v=\chi\nabla c$
        & periodic $x,y,z$ \\
        Porous convection
        & $C$
        & $\bm u=-\nabla p+C\bm e_z$, $\nabla\!\cdot\bm u=0$
        & Darcy velocity $\bm u$
        & periodic $x,y$; bounded $z$ \\
        \bottomrule
    \end{tabular}}
\end{table}

\subsection{Equations and coefficients}

\paragraph{Kolmogorov flow.}
On $[0,2\pi)^2$, the reference trajectories solve
\begin{equation}
    \partial_t\omega+\bm u\cdot\nabla\omega
    =\nu\Delta\omega-\alpha\omega-4\cos(4x_2),
    \qquad
    \Delta\psi=\omega-\bar\omega,
    \quad \bm u=\nabla^\perp\psi,
\end{equation}
with $\nu=10^{-2}$, $\alpha=0.1$, and frame spacing $\Delta t=0.1$. The pseudo-spectral reference solver uses $2/3$ de-aliasing, ETDRK2~\cite{cox2002etd}, and 20 internal steps per stored frame. Initial vorticity is obtained by Fourier filtering independent standard Gaussian grid values to $\|\bm k\|_\infty\leq8$, removing the zero mode, and normalizing the root-mean-square amplitude to one. The field is advanced for 500 stored-frame intervals before collection.

\paragraph{Dissipative surface quasi-geostrophy.}
On the same periodic domain,
\begin{equation}
    \partial_t\theta+\bm u\cdot\nabla\theta
    =\kappa\Delta\theta-\gamma(-\Delta)^{-1}\theta+f,
    \qquad
    \psi=-(-\Delta)^{-1/2}\theta,
    \quad \bm u=\nabla^\perp\psi,
\end{equation}
with $\kappa=0.008$, $\gamma=0.15$, and $\Delta t=0.1$. Each trajectory has a frozen random forcing supported on $3\leq|\bm k|\leq4$ with RMS amplitude $0.10$. The de-aliased ETDRK2 solver uses 40 internal steps per stored frame. Initial conditions use the same Gaussian filtering procedure as Kolmogorov flow, with $\|\bm k\|_\infty\leq8$, root-mean-square amplitude $0.5$, and a 200-frame warm-up.

\paragraph{Vlasov--Poisson.}
The phase-space domain is $x\in[0,4\pi)$ and $v\in[-6,6)$, with
\begin{equation}
    \partial_t f+v\,\partial_xf-E\,\partial_vf=0,
    \qquad \rho(x)=\int_{-6}^{6} f(x,v)\,\mathrm dv,
    \qquad \partial_xE=1-\rho,
    \label{eq:vp_full}
\end{equation}
and $\Delta t=0.1$.  A spectral Strang solver~\cite{strang1968splitting,cheng1976vlasov} applies a half $x$ translation, a full $v$ translation, and a second half $x$ translation, using five internal steps per stored frame.  The reference generator implements these translations by FFT on the truncated velocity interval; GRNO instead assigns zero to routed reads that leave $[-6,6)$.  The sampled initial distributions have negligible support at that boundary. Each trajectory independently selects a Landau or two-stream initial condition with probability $1/2$. Writing
$M_\sigma(v)=\exp[-v^2/(2\sigma^2)]/(\sqrt{2\pi}\sigma)$,
the unnormalized fields are
\begin{align}
 f_{\mathrm L}(x,v)
 &=M_\sigma(v)\left[1+\alpha\cos(x/2+\phi)
       +\tfrac14\alpha\cos(x-2\phi)\right],\\
 f_{\mathrm T}(x,v)
 &=\tfrac12\left[M_\sigma(v-v_0)+M_\sigma(v+v_0)\right]
       \left[1+\alpha\cos(x/2+\phi)
       +\tfrac15\alpha\cos(3x/2-\phi)\right].
\end{align}
Parameters are sampled independently and uniformly: $\phi\in[0,2\pi)$; $\sigma\in[0.8,1.2]$ and $\alpha\in[0.05,0.18]$ for Landau; $v_0\in[1.6,2.4]$, $\sigma\in[0.35,0.60]$, and $\alpha\in[0.03,0.10]$ for two-stream. Each field is divided by its discrete mean density, $(\Delta x\Delta v/L_x)\sum_{i,j}f(x_i,v_j)$, so that the spatially averaged density is one.

\paragraph{Screened Keller--Segel.}
On $[0,2\pi)^3$,
\begin{equation}
    \partial_t\rho=D\Delta\rho-\nabla\!\cdot(\rho\bm v),
    \qquad \bm v=\chi\nabla c,
    \qquad (-\Delta+\lambda^2)c=\rho-\bar\rho,
    \label{eq:keller_full}
\end{equation}
with $D=0.10$, $\chi=0.30$, $\lambda=1$, and $\Delta t=0.25$. The $2/3$-de-aliased Fourier ETD2 solver uses eight internal steps per stored frame. Initial densities have the form $\rho=\exp(0.2\xi)$, where $\xi$ is a unit-variance random field supported on $1\leq|\bm k|\leq3$; each sample is rescaled to mean one.

\paragraph{Buoyancy-driven porous convection.}
On the unit cube, with periodic horizontal coordinates and bounded vertical coordinate $z$,
\begin{equation}
    \partial_t C+\nabla\!\cdot(\bm u C)=Ra_D^{-1}\Delta C,
    \qquad \bm u=-\nabla p+C\bm e_z,
    \qquad \nabla\!\cdot\bm u=0,
    \label{eq:porous_full}
\end{equation}
with $Ra_D=1000$ and $\Delta t=0.1$.  We impose $C=1$ at the top, $\partial_zC=0$ at the bottom, and impermeability $u_z=0$ at both vertical boundaries.  Pressure is fixed to zero mean.  The finite-volume reference solver combines MUSCL advection, SSP/RK2 time stepping, an exact mixed-boundary diffusion semigroup, and Strang splitting; its internal step is at most $0.02$ and also satisfies an advective CFL bound of $0.35$. Trajectories are generated on $128^3$ and volume-averaged to $64^3$. The vertical coordinate increases downward from the top boundary, $z=0$. Before the warm-up, concentration is initialized as
\begin{equation}
 C_0(x,y,z)=\operatorname{clip}_{[0,1]}
 \left[b(z)+0.08b(z)(1-b(z))\xi(x,y)\right],
 \qquad
 b(z)=\operatorname{erfc}\!\left(
 \frac{z}{2\sqrt{Ra_D^{-1}t_0}}\right).
\end{equation}
Here $t_0$ is sampled uniformly from $[0.9,1.1]$. The horizontal field $\xi$ is a sum of sine and cosine modes with integer wavevectors $1\leq|\bm k|\leq8$, using one representative of each pair $\{\bm k,-\bm k\}$. Its coefficients are independent standard Gaussian variables, and the field is normalized to unit root-mean-square amplitude. The initialized field is advanced for $0.2$ before storage.

\subsection{Dataset contract}

\begin{table}[t]
    \centering
    \caption{\textbf{Dataset contract.}  Shapes list trajectories $\times$
    stored frames; every frame also contains the grid shown in the second
    column.  Splits are fixed before training.}
    \label{tab:data_contract}
    \small
    \setlength{\tabcolsep}{4.5pt}
    \resizebox{\textwidth}{!}{%
    \begin{tabular}{lccccc}
        \toprule
        System & Grid & Train & Validation & Test & Stored $\Delta t$ \\
        \midrule
        Kolmogorov & $128^2$ & $64\times51$ & $8\times101$ & $12\times101$ & $0.10$ \\
        SQG & $128^2$ & $64\times51$ & $8\times101$ & $12\times101$ & $0.10$ \\
        Vlasov--Poisson & $128_x\times128_v$ & $64\times101$ & $8\times121$ & $12\times161$ & $0.10$ \\
        Keller--Segel & $64^3$ & $64\times51$ & $8\times101$ & $12\times101$ & $0.25$ \\
        Porous convection & $64^3$ & $64\times101$ & $8\times101$ & $12\times101$ & $0.10$ \\
        \bottomrule
    \end{tabular}}
\end{table}

Each training run records a trajectory/start-time manifest; a result is included in the matched table only after its manifest hash agrees with the task reference. Manifests log array shapes, coefficients, split membership, solver version, and file hashes; any run with a mismatched contract is excluded.

\section{Equation adapter, traces, and boundaries}
\label{app:trace_details}
The adapter is deterministic, has no trainable parameters, and is recomputed from the current predicted state at every rollout step.

\subsection{Diagnostic solves}

\paragraph{Periodic active scalars.}
For Kolmogorov flow and SQG, the zero Fourier mode is fixed to zero and
\begin{equation}
    \widehat\psi(\bm k)=-|\bm k|^{-r}\widehat q(\bm k),
    \qquad \widehat{\bm v}(\bm k)
      =(ik_2,-ik_1)\widehat\psi(\bm k),
    \label{eq:active_scalar_adapter}
\end{equation}
with $r=2$ for vorticity and $r=1$ for SQG. This is the discrete Green multiplier for the respective periodic diagnostic relation.

\paragraph{Vlasov--Poisson.}
The adapter integrates $f$ over velocity, removes the neutral Fourier mode, and solves Eq.~(\ref{eq:vp_full}) spectrally for $E(x)$.  It returns the phase-space carrier $(v,-E(x))$ rather than treating $E$ as a second evolved state.

\paragraph{Keller--Segel.}
The screened inverse is diagonal in Fourier space:
\begin{equation}
    \widehat c(\bm k)
       =\frac{\widehat\rho(\bm k)}{|\bm k|^2+\lambda^2}
       \quad(\bm k\neq0),
    \qquad \bm v=\chi\nabla c .
\end{equation}
The mean source is removed as in Eq.~(\ref{eq:keller_full}).

\paragraph{Porous convection.}
The adapter first forms the buoyancy on vertical cell faces: interior faces use the average of their two adjacent concentrations, and the top and bottom face values are zero. The discrete divergence of this face buoyancy is the pressure right-hand side. The pressure equation uses the finite-volume Laplacian, diagonalized by Fourier transforms in the periodic horizontal directions and a cosine transform vertically; the zero mode is set to zero. Face velocities are computed from this buoyancy and the corresponding discrete pressure gradient, with zero normal velocity at the top and bottom. Averaging adjacent face velocities gives the cell-centered carrier supplied to the network.

\subsection{Backward traces}

For spatial transport, GRNO uses the midpoint rule from Eq.~(\ref{eq:midpoint_route}): the adapter first samples the carrier at $\bm x-\tfrac{1}{2}\Delta t\,\bm v(\bm x)$ and then places the final read at $\bm x-\Delta t\,\bm v_{1/2}(\bm x)$.

Vlasov--Poisson uses the symmetric phase-space trace matching the order of the characteristics:
\begin{equation}
\begin{aligned}
    x_{1/2}&=x-\tfrac12\Delta t\,v,\\
    E_{1/2}&=\mathcal I[E](x_{1/2}),\\
    v_d&=v+\Delta t\,E_{1/2},\\
    x_d&=x_{1/2}-\tfrac12\Delta t\,v_d.
\end{aligned}
\label{eq:vp_trace}
\end{equation}
The routed read is taken at $(x_d,v_d)$.  The sign follows $\dot x=v$ and $\dot v=-E$ in Eq.~(\ref{eq:vp_full}) while integrating backward from the target point.

\subsection{Interpolation and boundary rules}

\begin{table}[h]
    \centering
    \caption{\textbf{Boundary behavior of routed interpolation.}  The rule is
    applied identically at every latent scale.}
    \label{tab:boundary_rules}
    \small
    \begin{tabular}{lll}
        \toprule
        Coordinate type & Coordinate handling & Sample outside domain \\
        \midrule
        Periodic & wrap modulo domain length & wrapped value \\
        Bounded wall & clamp to the nearest grid index & value at the nearest boundary grid point \\
        Phase-space outflow & mask each out-of-range index & zero contribution \\
        \bottomrule
    \end{tabular}
\end{table}

Bilinear interpolation is used in two dimensions and trilinear interpolation in three dimensions. At scale $\ell$, a departure displacement $d_{\ell,j}$ shifts the fractional grid index by $d_{\ell,j}/\Delta x_{\ell,j}$, where $\Delta x_{\ell,j}=L_j/N_{\ell,j}$. The full-resolution displacement is average-pooled to each scale and remains in physical units. Each neighboring index in the interpolation stencil is wrapped, clamped, or assigned zero interpolation weight according to Table~\ref{tab:boundary_rules}.

\section{Shared GRNO architecture}
\label{app:architecture}

The trainable core uses a common 2D architecture and its direct 3D counterpart. Within a dimension, the parameter names, tensor shapes, stage depths, and channel widths are identical across PDEs. Only the equation adapter, boundary metadata, context channel, and deterministic output constraint vary by task.

\begin{table}[t]
    \centering
    \caption{\textbf{Dimension-specific GRNO configuration.}  ``Axial'' in
    3D denotes the average of three depthwise kernels, one per spatial axis.}
    \label{tab:grno_architecture}
    \small
    \setlength{\tabcolsep}{5pt}
    \begin{tabular}{lcc}
        \toprule
        Quantity & 2D / phase space & 3D \\
        \midrule
        Input channels & 5 & 6 \\
        Base width & 48 & 24 \\
        Stage widths & $(48,96,144,192)$ & $(24,48,72,96)$ \\
        Encoder depths & $(2,2,3,4)$ & $(2,2,3,4)$ \\
        Decoder blocks per scale & 1 & 1 \\
        Local depthwise kernel & $5\times5$ & axial $5$-tap \\
        Number of scales & 4 & 4 \\
        Trainable parameters & $1{,}534{,}945$ & $391{,}561$ \\
        Trainable adapter parameters & 0 & 0 \\
        \bottomrule
    \end{tabular}
\end{table}

\paragraph{Inputs and normalization.}
Each task advances one scalar channel. The neural input has shape $B\times5\times N_y\times N_x$ in two dimensions and $B\times6\times N_z\times N_y\times N_x$ in three dimensions. For Vlasov--Poisson the two grid axes are $(v,x)$. The channel order is state, carrier components in $(x,y)$ or $(x,y,z)$ order, one context channel, and $\Delta t/\Delta t_\mu$. Using fixed scales computed from training data, 
\begin{equation}
 \mathcal N_{q,\mu}(q)=
 \begin{cases}
 q/s_{q,\mu}, & \text{2D and phase space},\\
 (q-m_{q,\mu})/s_{q,\mu}, & \text{3D},
 \end{cases}
 \qquad
 [\mathcal N_{v,\mu}(\bm v)]_j=v_j/s_{v,j,\mu}.
\end{equation}
Input scale denominators are floored at $10^{-7}$. For Kolmogorov flow and SQG, state, velocity, forcing and increment scales are read from the training normalization file; forcing is divided by its scale, and the context channel is the normalized forcing pattern. For Vlasov--Poisson, state and increment scales are root-mean-square values estimated from up to 512 training frame pairs using sampling seed 17. Its carrier scales are $v_{\max}$ and the root-mean-square electric field, and its context is zero. For the 3D tasks, state mean, state scale and increment scale are read from the saved training statistics. Each carrier component is scaled by its root-mean-square value over 24 evenly selected training states. Porous convection uses $c_\mu(z)=2z/L_z-1$ as context; Keller--Segel uses zero.

\paragraph{Multiscale fusion.}
The grids have side lengths $(128,64,32,16)$ in 2D and $(64,32,16,8)$ in 3D. Every encoder scale produces one feature tensor $s_\ell$ and its routed copy $r_\ell$. The coarsest decoder input is $[s_3,r_3]$; at the remaining scales,
\begin{equation}
 z_\ell=\mathrm{Dec}_\ell\!\left(
 W_\ell[\mathrm{Up}_\ell(z_{\ell+1}),s_\ell,r_\ell]
 \right),\qquad \ell=2,1,0.
\end{equation}
Brackets denote channel concatenation, and $W_\ell$ is a learned pointwise projection. Downsampling uses average pooling and a pointwise projection. Upsampling uses a pointwise projection followed by nearest-neighbor interpolation. A local residual block applies
\begin{equation}
 u\longmapsto u+W_2\operatorname{GELU}
 \bigl(W_1\operatorname{LN}(\operatorname{DW}(u))\bigr).
\end{equation}
Here $\operatorname{DW}$ is a depthwise spatial convolution, with the three axial outputs averaged in 3D. Layer normalization acts across channels independently at each grid point, with learned scale and shift and numerical floor $10^{-6}$. The pointwise maps expand channels by a factor of two and project them back.

The increment head $H_\theta$ is a learned pointwise projection from the finest decoder feature $z_0$ to the normalized state increment and is initialized to zero. For tasks with an exactly conserved spatial integral, the predicted increment is projected to zero mean. Porous convection uses the identity projection because mass enters through its Dirichlet top boundary. These projections are fixed and contain no learned coefficients.

\section{Training and evaluation contract}
\label{app:protocol}

\subsection{Optimization and exposure}

\begin{table}[t]
    \centering
    \caption{\textbf{Matched training and evaluation protocol.} Four 3D micro-batches are accumulated to obtain an effective batch size of four.}
    \label{tab:training_contract}
    \small
    \setlength{\tabcolsep}{4pt}
    \resizebox{\textwidth}{!}{%
    \begin{tabular}{lcc}
        \toprule
        Setting & 2D / phase space & 3D \\
        \midrule
        Epochs & 100 & 100 \\
        Windows per epoch & 2,048 & 512 \\
        Micro-batch / accumulation & $4/1$ & $1/4$ \\
        Effective batch & 4 & 4 \\
        Optimizer updates per epoch & 512 & 128 \\
        Training rollout $K$ & 4 & 4 \\
        Validation / test horizon & $10/40$ & $10/40$ \\
        Optimizer & AdamW & AdamW \\
        Initial learning rate & $3\times10^{-4}$ & $3\times10^{-4}$ \\
        Weight decay & $10^{-4}$ & $10^{-4}$ \\
        Schedule & cosine to $1.5\times10^{-5}$ & cosine to $1.5\times10^{-5}$ \\
        Gradient clipping & 1.0 & 1.0 \\
        Precision & bfloat16 AMP & bfloat16 AMP \\
        Seeds: baseline comparison & $\{0,1,2\}$ & $\{0,1,2\}$ \\
        Seeds: fixed-weight route interventions & $\{0,1,2\}$ & $\{0,1,2\}$ \\ Seed: independently trained ablations & $0$ & $0$ \\
        \bottomrule
    \end{tabular}}
\end{table}

For a fixed task, all models receive the same window manifest.  An epoch is therefore defined by data exposure and optimizer updates, not by how many micro-batches fit in memory.  Three-dimensional models use gradient accumulation only to meet the A100 memory budget.  Training uses AdamW~\cite{loshchilov2019adamw}; the learning-rate scheduler advances once per epoch. For validation, all admissible (trajectory, start-frame) pairs are ordered by trajectory and then start frame. The evaluator selects 64 evenly spaced pairs in 2D and 32 in 3D, or all pairs if fewer are available. Checkpoint selection minimizes relative $L^2$ averaged over these windows and all ten validation horizons.

\subsection{Loss}

Let $e_b^{(k)}=\widehat q_b^{\,n+k}-q_b^{n+k}$ for micro-batch sample $b$. Let $B$ be the micro-batch size, $N$ the number of grid points, and $s_{q,\mu}$ the fixed state scale defined in Appendix~\ref{app:architecture}. All reported tasks have one physical output channel. The state loss is 
\begin{equation}
 \ell_{\mathrm{state}}^{(k)}=
 \frac{1}{BN}\sum_{b=1}^{B}\sum_{\bm i}
 \left|e_b^{(k)}(\bm i)/s_{q,\mu}\right|^2.
\end{equation}
Define $\langle a\rangle=(BN)^{-1}\sum_{b,\bm i}a_b(\bm i)$. The gradient loss is
\begin{equation}
 \ell_\nabla^{(k)}=
 \frac{\sum_j\langle|\delta_j e^{(k)}|^2\rangle}
 {\max\{\sum_j\langle|\delta_j q^{n+k}|^2\rangle,\varepsilon\}}.
\end{equation}
The difference $\delta_j$ uses grid spacing $L_j/N_j$: centered differences with wraparound on periodic axes, centered interior differences and one-sided endpoint differences otherwise. Let $\mathcal F_{\mathcal P}$ transform only the periodic axes. The spectral loss averages per-sample ratios:
\begin{equation}
 \ell_{\mathrm{spec}}^{(k)}=
 \frac1B\sum_{b=1}^{B}
 \frac{\bigl\||\mathcal F_{\mathcal P}\widehat q_b^{\,n+k}|
                  -|\mathcal F_{\mathcal P}q_b^{n+k}|\bigr\|_2^2}
 {\max\{\|\mathcal F_{\mathcal P}q_b^{n+k}\|_2^2,\varepsilon\}}.
\end{equation}
The spectral norm sums over every transformed coefficient and any remaining grid axis. The gradient and spectral terms use physical field values. The implementation uses $\varepsilon=10^{-8}$, $\lambda_\nabla=0.05$, $\lambda_{\mathrm{spec}}=0.02$, $\rho=0.8$, and $Z_K=\sum_{k=1}^{K}\rho^{k-1}$ in Eq.~(\ref{eq:grno_training}). In 3D, gradients from four micro-batches are accumulated for each optimizer update. Gradients pass through all four rollout steps, including the recomputed diagnostic solves, traces, and interpolations.

\subsection{Free autoregressive evaluation}

For test trajectory $i\in\{0,\ldots,11\}$, let $s_i$ be the prescribed start frame. Evaluation uses
\begin{equation}
 \widehat q_i^0=q_i^{s_i},\qquad
 \widehat q_i^{h+1}=\mathcal M_\theta(\widehat q_i^h),
 \qquad h=0,\ldots,39,
\end{equation}
and compares $\widehat q_i^h$ with $q_i^{s_i+h}$. The three 2D systems use $s_i=0$. Porous convection uses $s_i=20(i\bmod4)$. Keller--Segel uses the integer start indices $(0,5,10,16,21,27,32,38,43,49,54,60)$, obtained by evenly spacing twelve indices over the admissible interval $[0,60]$. Errors are averaged over the twelve windows within each seed. Tables~\ref{tab:main_results} and \ref{tab:route_interventions} report the mean and sample standard deviation across the three seed means. Table~\ref{tab:factorial_five_pde_seed0} reports seed~0.

\section{Baseline provenance and matched adaptations}
\label{app:baselines}

\begin{table}[t]
    \centering
    \caption{\textbf{Baseline roles and adaptations.} For dimensional or autoregressive adaptations, we retain each published model's interaction mechanism and modify only the input and output interfaces required by the common one-frame protocol. The table summarizes the evaluated implementations.}
    \label{tab:baseline_provenance}
    \scriptsize
    \setlength{\tabcolsep}{4pt}
    \begin{tabularx}{\textwidth}{l>{\raggedright\arraybackslash}p{0.27\textwidth}
                                  >{\raggedright\arraybackslash}X}
        \toprule
        Model & Retained interaction mechanism & Protocol adaptation \\
        \midrule
        ReViT & canonical local frames and rotationally equivariant
          transformer blocks
        & Physical-feature adapter constructs the vector/scalar layout from
          each task; one-frame residual decoder. \\
         Flowers & learned multihead coordinate warps and latent resampling
        & The same state, context, boundary sampler, and autoregressive evaluator
          are used; read coordinates are learned from forecast data. \\
        NIPS-AR & channel-independent Fourier kernel and linear attention
        & Released spatial block placed in a recurrent one-frame wrapper;
          reported as an autoregressive adaptation. \\
        RieszNO / RNO-3D & Riesz directional representation plus spectral
          mixing
        & Released 2D construction; direct axial 3D extension is explicitly
          marked as a dimensional adaptation. \\
        \bottomrule
    \end{tabularx}
\end{table}

For every baseline reported in the matched comparison, the implementation uses the model-specific adaptation listed in Table~\ref{tab:baseline_provenance}. Its checkpoint and manifest are verified against the common array files, sampled windows, training-only statistics, $K=4$ loss, optimizer schedule, effective batch, update count, validation selection, and $H=40$ evaluator. Micro-batch size is adjusted only for memory.

Each experiment manifest records the source repository revision, architecture configuration, parameter count, peak allocated memory, and forward time for that run.

\section{Metrics and result records}
\label{app:metrics}

For each test window, errors are evaluated in physical units. At rollout step $h$, the relative $L^2$ error is
\begin{equation}
 E_{L^2}(h)=
 \frac{\|\widehat q_h-q_h\|_2}{\max\{\|q_h\|_2,\varepsilon\}},
 \qquad \varepsilon=10^{-8}.
\end{equation}
The rollout mean is $H^{-1}\sum_{h=1}^{H}E_{L^2}(h)$. The reported spectral error is
\begin{equation}
 E_{\mathrm{spec}}(h)=
 \left[
 \frac{\bigl\||\mathcal F_{\mathcal P}\widehat q_h|
                  -|\mathcal F_{\mathcal P}q_h|\bigr\|_2^2}
 {\max\{\|\mathcal F_{\mathcal P}q_h\|_2^2,\varepsilon\}}
 \right]^{1/2}.
\end{equation}
Thus the per-window spectral metric is the square root of the spectral ratio used in training. Relative $H^1$ adds the discrete derivative energies to the squared field norms in numerator and denominator before taking the square root. Spatial correlation is computed after subtracting each field's spatial mean. Persistence uses $\widehat q_h=q_0$, and its skill score is
\begin{equation}
 \mathrm{Skill}(h)=1-
 \frac{E_{L^2}^{\mathrm{model}}(h)}
 {\max\{E_{L^2}^{\mathrm{persistence}}(h),\varepsilon\}}.
\end{equation}

\begin{table}[h]
    \centering
    \caption{\textbf{Task-specific physical metrics recorded by the common evaluator.} These metrics are reported in addition to relative $L^2$ error.}
    \label{tab:physical_metrics}
    \small
    \begin{tabularx}{\textwidth}{lX}
        \toprule
        Task & Additional records \\
        \midrule
        Kolmogorov / SQG
        & correlation, spectral error, variance error, and kinetic-energy or
          scalar-variance error \\
        Vlasov--Poisson
        & total-mass drift, negative-density fraction, charge-density error,
          and electric-energy error \\
        Keller--Segel
        & mass drift, negative-mass fraction, peak and variance error, and
          screened free-energy error \\
        Porous convection
        & anomaly error, plume-mask IoU, top dissolution-flux error,
          plume-depth error, scalar-dissipation error, variance error, and
          fractions below zero or above one \\
        \bottomrule
    \end{tabularx}
\end{table}

Errors are computed for each trajectory and rollout step. Within each seed, results are averaged over the 12 prescribed test windows. Tables~\ref{tab:main_results} and \ref{tab:route_interventions} report the mean and sample standard deviation across the three seed means, whereas Table~\ref{tab:factorial_five_pde_seed0} reports seed~0.

\section{Expanded route-mechanism study}
\label{app:route_study}

\subsection{What each intervention changes}

\begin{table}[h]
    \centering
    \caption{\textbf{Fixed-weight route interventions.} Network weights, construction of carrier features, centered branches, and prescribed test windows are held fixed.}
    \label{tab:intervention_semantics}
    \small
    \begin{tabularx}{\textwidth}{lXX}
        \toprule
        Mode & Operation on displacement $\bm d$ & Question tested \\
        \midrule
        Correct & $d$ from the equation adapter & reference computation \\
        Zero & $d\leftarrow0$ & Is the carrier used as a coordinate rather
          than only as an input feature? \\
        Reverse & $d\leftarrow-d$ & Does route orientation matter? \\
        Shift & deterministic spatial roll of $d$ & Does alignment between
          route and current state matter? \\
        \bottomrule
    \end{tabularx}
\end{table}

At each rollout step, the carrier and departure displacement are recomputed from that run's current predicted state. The chosen intervention is applied to this new displacement before pooling to each feature scale and sampling the features. For the \emph{Shift} intervention, the fixed offsets are $(\lfloor N_y/3\rfloor,\lfloor N_x/5\rfloor)$ in 2D and $(\lfloor N_z/5\rfloor,\lfloor N_y/4\rfloor,\lfloor N_x/3\rfloor)$ in 3D. The displacement values continue to evolve with the predicted state at every rollout step.

Final intervention errors are reported in Table~\ref{tab:route_interventions}; Figure~\ref{fig:rollout_interventions}c--d shows the corresponding error ratios and rollout profiles.

\subsection{Independently trained controls}
\label{app:trained_controls}

All variants share the core architecture within each dimension. Disabling carrier features sets their input channels to zero; disabling routing sets the latent displacement to zero. The centered and routed branches remain in place, giving the four factorial variants identical parameter counts. The \emph{Shifted equation-specified route} control applies the same deterministic shift as above throughout both training and testing.

\paragraph{Learned route.}
The learned-route displacement head has a pointwise input projection, one local residual block, and a pointwise output projection. Its width is 24 in 2D and 12 in 3D; it outputs two or three displacement components. The final projection is initialized to zero. For each component,
\begin{equation}
 d_{\phi,j}(q)=a_j\tanh\!\left([H_\phi(x_\mu(q))]_j\right),
 \qquad
 a_j=\max\{3\,\operatorname{RMS}_{\mathrm{train}}(d_j),\Delta x_j\}.
\end{equation}
Here $d_j$ is the equation-specified departure displacement on training states and $\Delta x_j$ is the full-grid spacing. The 2D scale estimate uses 64 training states sampled with replacement using seed 23071; the 3D estimate uses 16 evenly spaced entries in the ordered list of training trajectory/frame pairs. The head receives the same normalized input as the forecasting core and is optimized jointly with it through the forecasting loss.

\paragraph{State advection + residual.}
The \emph{State advection + residual} control first computes $q_{\mathrm{adv}}=\mathcal I[q](\bm\Xi[q])$ and then applies
\begin{equation}
 \widehat q^+=q+\Pi_\mu\!\left(
 q_{\mathrm{adv}}-q+
 \sigma_{\Delta q,\mu}R_\theta(x_\mu(q))\right).
\end{equation}
The centered network receives the current state, carrier, and context, and all latent displacements are set to zero. It is trained from scratch with the same loss, training budget, and checkpoint rule. The task projection acts on the complete state increment, including $q_{\mathrm{adv}}-q$. The diagnostic carrier and departure map are recomputed from the current prediction at every step.

\subsection{Visualization sample and transect}

For Figure~\ref{fig:rollout_interventions}e, the selected test window maximizes the reduction in final relative $L^2$ error from \emph{Carrier feature only} to GRNO among the twelve prescribed windows. Within this window, the vertical line maximizes the reduction in absolute field error averaged over $h=1,\ldots,40$ and the vertical grid coordinate. Both independently trained models use the same window and line. The three concentration maps use a common color scale, and the same reference concentration contour is overlaid as a dashed line on each map.

\section{Additional Visualizations}
\label{app:additional_visualization}
\subsection{Additional qualitative forecasting results}
\label{app:qualitative_forecasting}
\begin{figure}[!htbp]
    \centering
    \includegraphics[width=\linewidth]{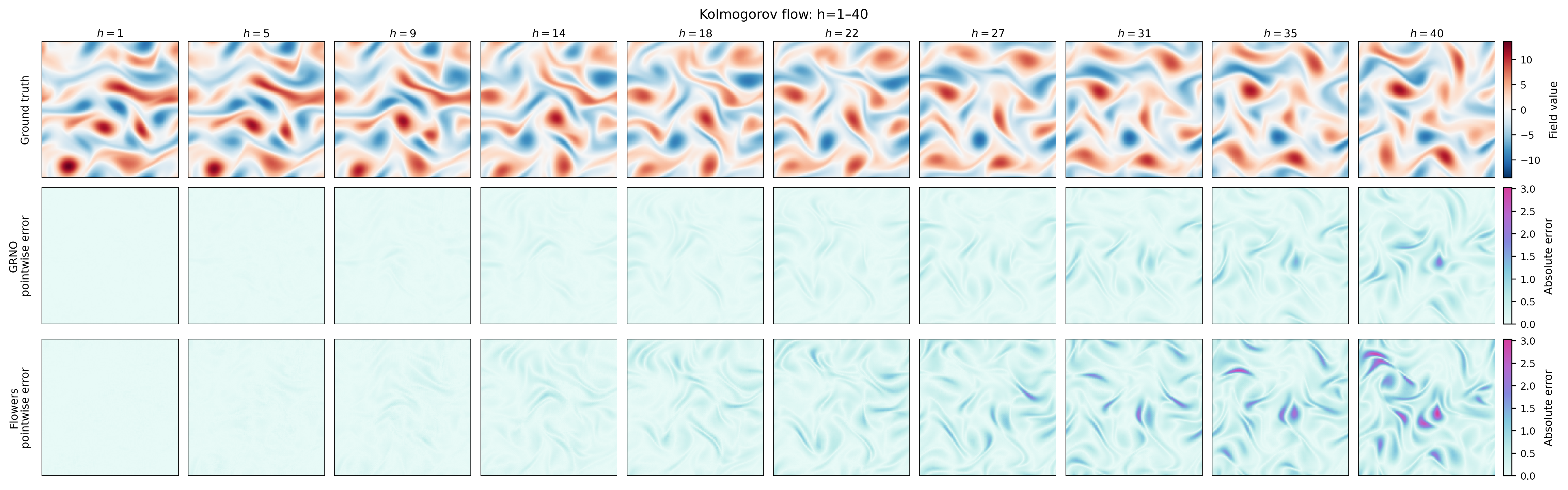}
    \caption{
    \textbf{Long-horizon qualitative comparison on Kolmogorov flow.}
    The first row shows the reference vorticity field at ten horizons spanning the 40-step free rollout.
    The second and third rows show the pointwise absolute errors of GRNO and Flowers, respectively.
    The error scale is shared by both models and all displayed horizons.
    }
    \label{fig:supp_qual_kolmogorov}
\end{figure}

\begin{figure}[!htbp]
    \centering
    \includegraphics[width=\linewidth]{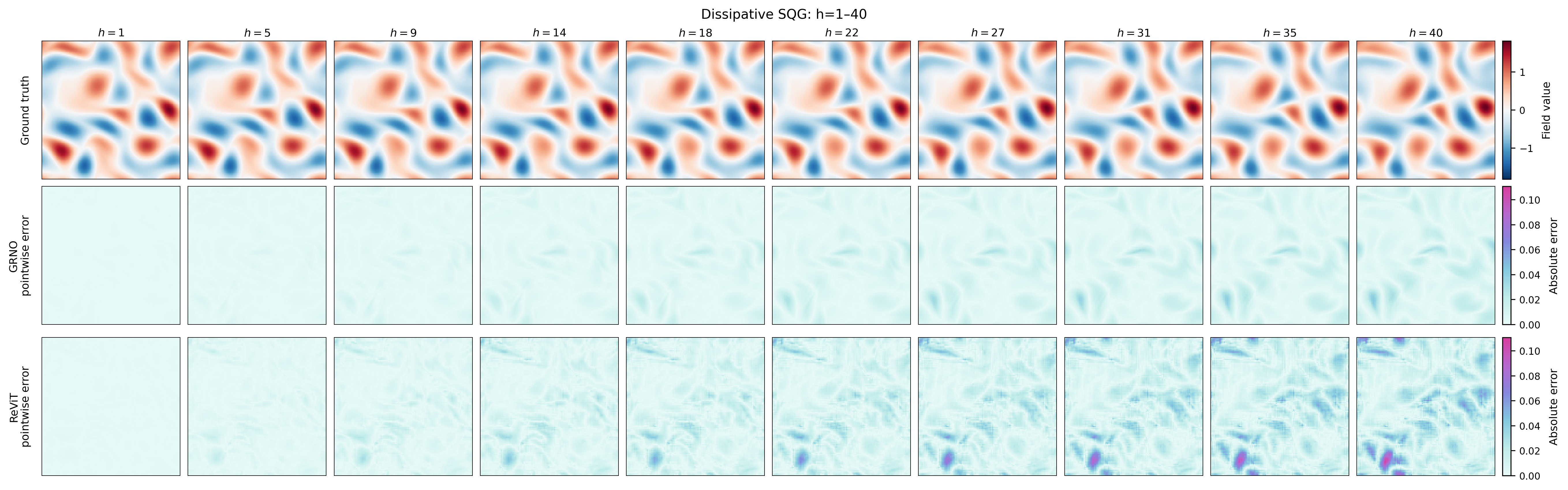}
    \caption{
    \textbf{Long-horizon qualitative comparison on dissipative SQG.}
    The first row shows the reference scalar field at ten horizons spanning the 40-step free rollout.
    The second and third rows show the pointwise absolute errors of GRNO and ReViT, respectively.
    The error scale is shared by both models and all displayed horizons.
    }
    \label{fig:supp_qual_sqg}
\end{figure}

\begin{figure}[!htbp]
    \centering
    \includegraphics[width=\linewidth]{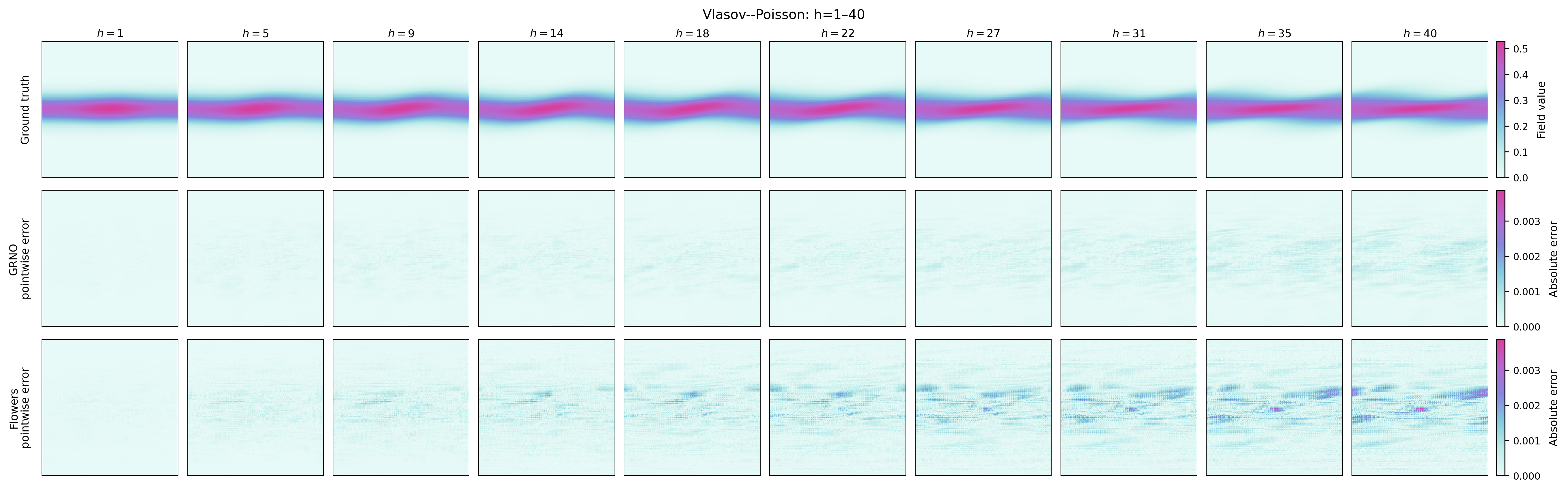}
    \caption{
    \textbf{Long-horizon qualitative comparison on Vlasov--Poisson.}
    The first row shows the reference phase-space distribution at ten horizons spanning the 40-step free rollout.
    The second and third rows show the pointwise absolute errors of GRNO and Flowers, respectively.
    The error scale is shared by both models and all displayed horizons.
    }
    \label{fig:supp_qual_vlasov_poisson}
\end{figure}

\begin{figure}[!htbp]
    \centering
    \includegraphics[width=\linewidth]{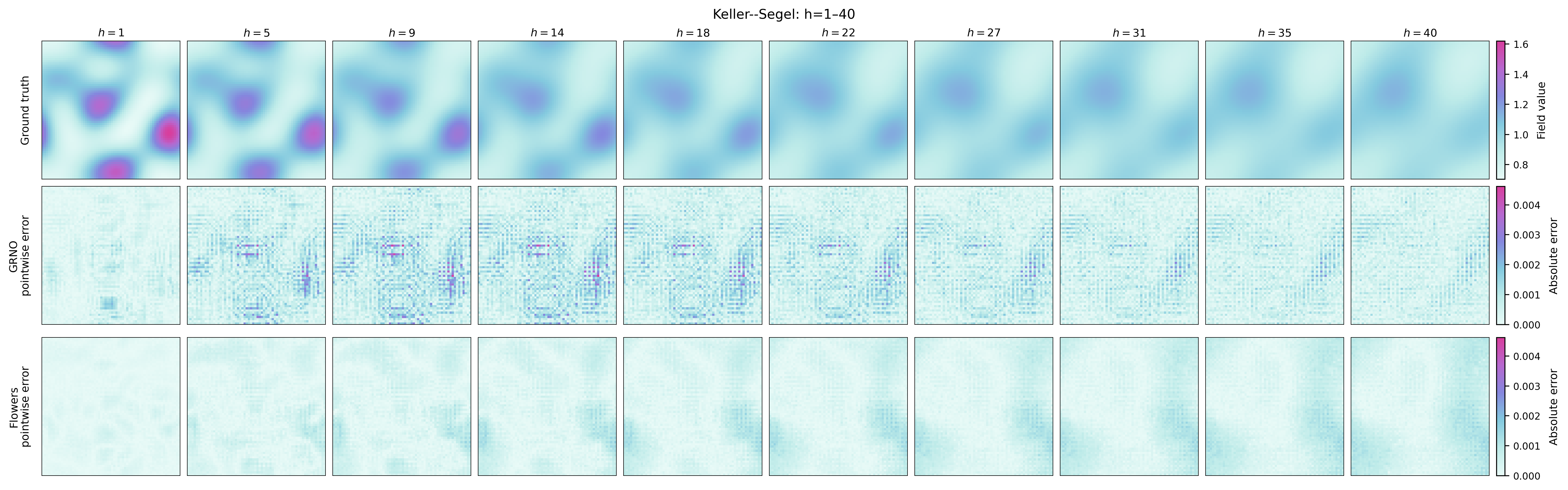}
    \caption{
    \textbf{Long-horizon qualitative comparison on Keller--Segel.}
    The first row shows the reference density field on the fixed visualization slice at ten horizons spanning the 40-step free rollout.
    The second and third rows show the pointwise absolute errors of GRNO and Flowers, respectively.
    The error scale is shared by both models and all displayed horizons.
    }
    \label{fig:supp_qual_keller_segel}
\end{figure}

\begin{figure}[!htbp]
    \centering
    \includegraphics[width=\linewidth]{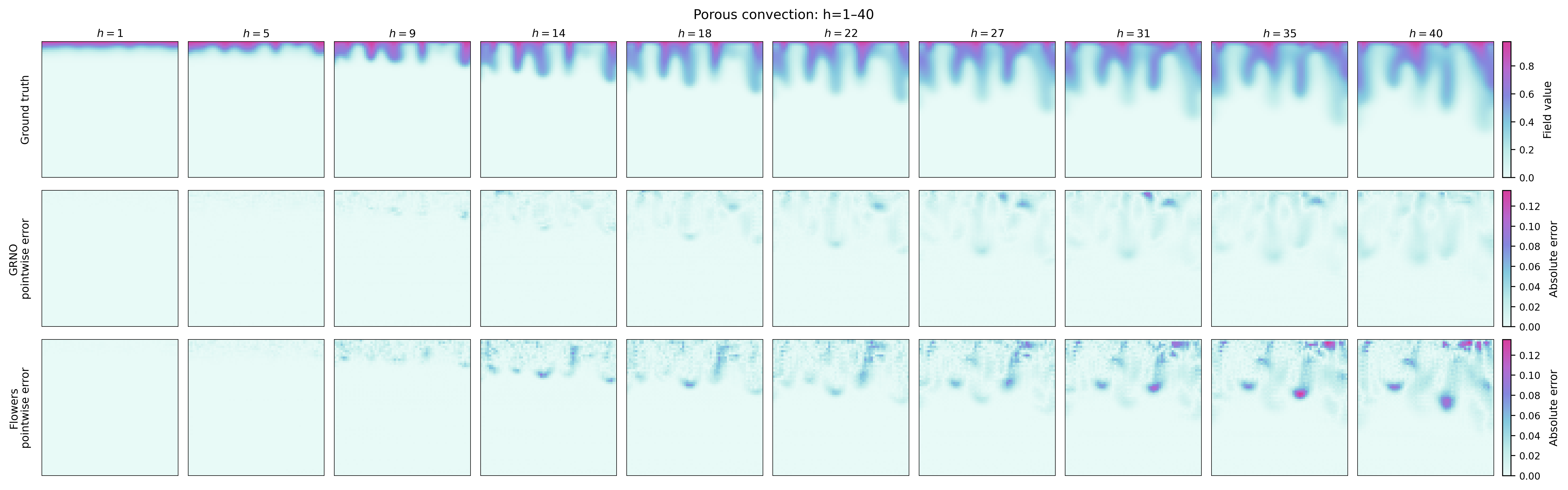}
    \caption{
    \textbf{Long-horizon qualitative comparison on porous convection.}
    The first row shows the reference concentration field on the fixed visualization slice at ten horizons spanning the 40-step free rollout.
    The second and third rows show the pointwise absolute errors of GRNO and Flowers, respectively.
    The error scale is shared by both models and all displayed horizons.
    }
    \label{fig:supp_qual_porous_convection}
\end{figure}

Figures~\ref{fig:supp_qual_kolmogorov}--\ref{fig:supp_qual_porous_convection} compare reference fields and pointwise absolute errors at ten horizons spanning the 40-step free rollout, using GRNO and the strongest baseline in Table~\ref{tab:main_results}.

\clearpage
\subsection{Qualitative effects of route interventions}
\label{app:qualitative_route_interventions}

\begin{figure}[!htbp]
    \centering
    \includegraphics[width=\linewidth]{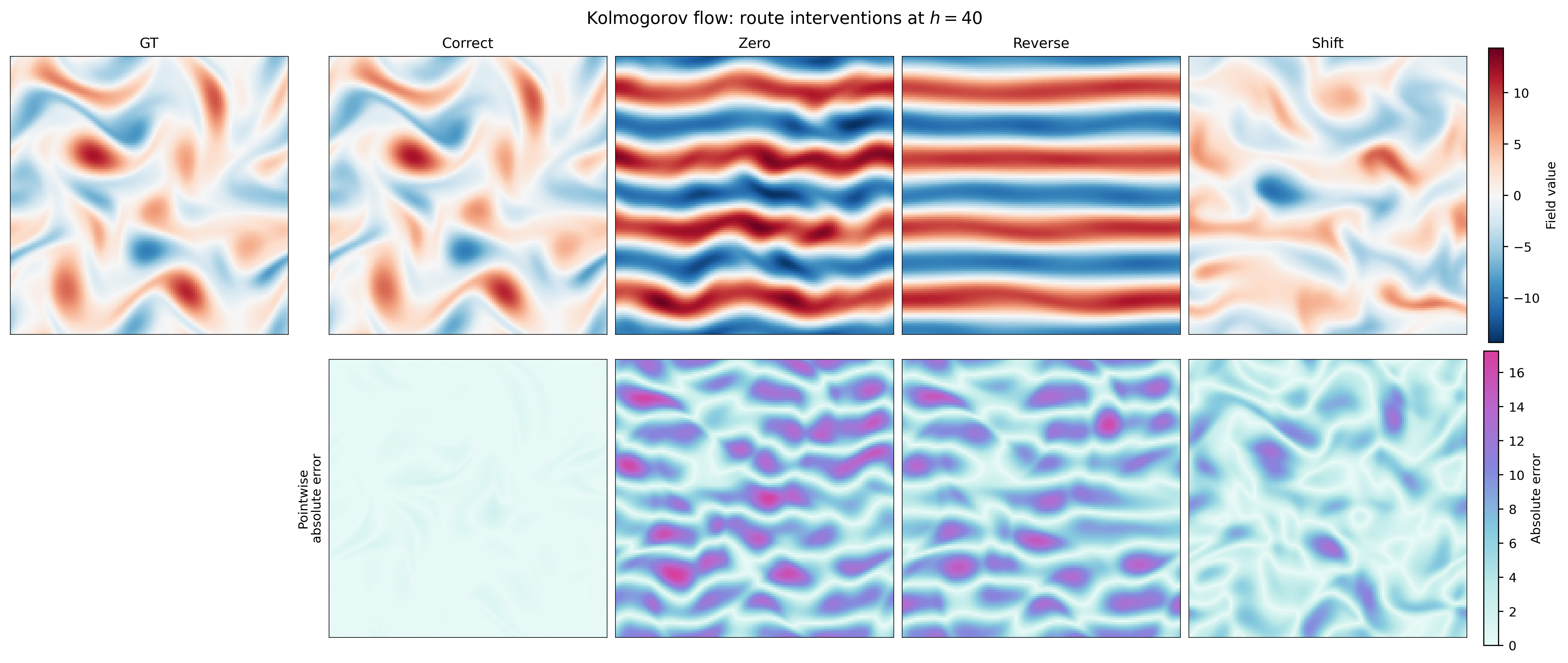}
    \caption{
    \textbf{Route interventions on Kolmogorov flow at $h=40$.}
    The top row shows the reference field and predictions using the
    correct, zero, reversed, and shifted routes.
    The bottom row shows the corresponding pointwise absolute errors.
    All predictions use the same fitted GRNO checkpoint; only the
    departure route is changed.
    }
    \label{fig:supp_route_kolmogorov}
\end{figure}

\begin{figure}[!htbp]
    \centering
    \includegraphics[width=\linewidth]{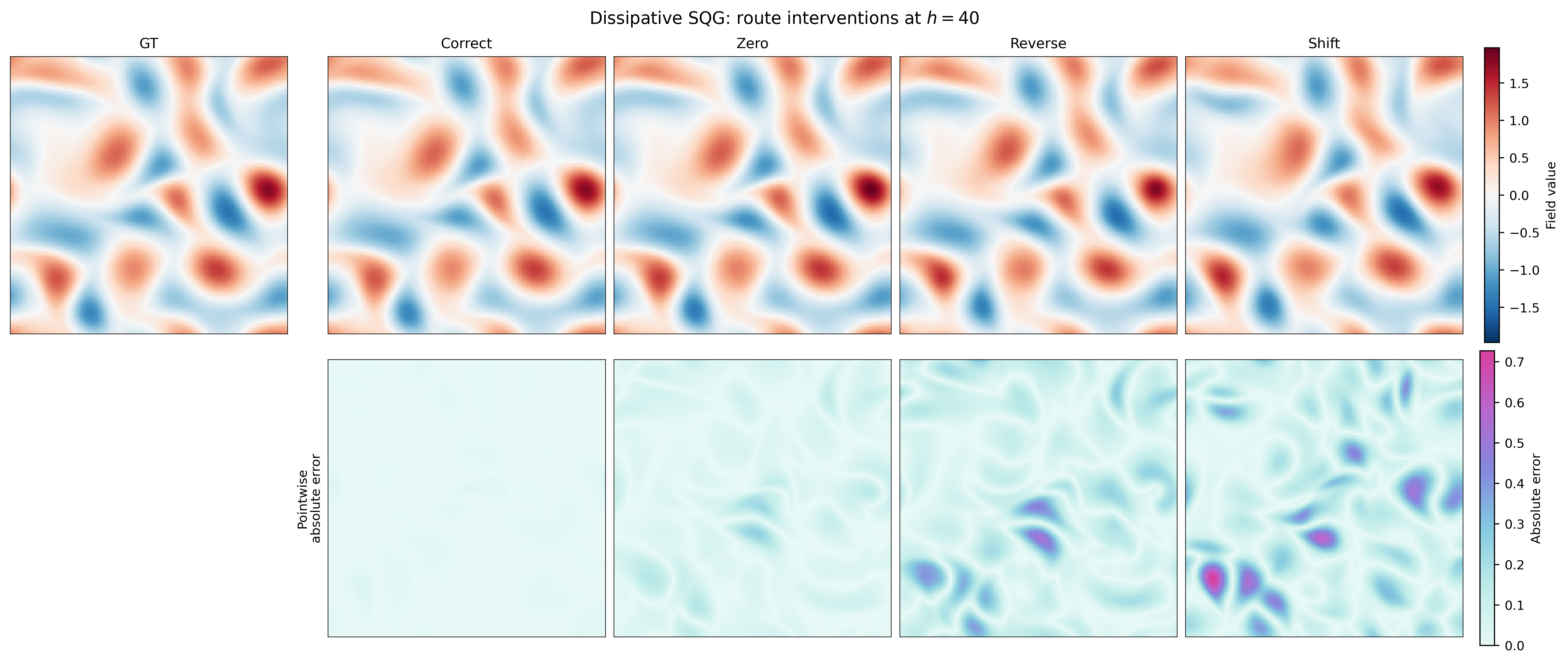}
    \caption{
    \textbf{Route interventions on dissipative SQG at $h=40$.} The top row shows the reference field and predictions using the correct, zero, reversed, and shifted routes. The bottom row shows the corresponding pointwise absolute errors. All predictions use the same fitted GRNO checkpoint; only the departure route is changed.
    }
    \label{fig:supp_route_sqg}
\end{figure}

\begin{figure}[!htbp]
    \centering
    \includegraphics[width=\linewidth]{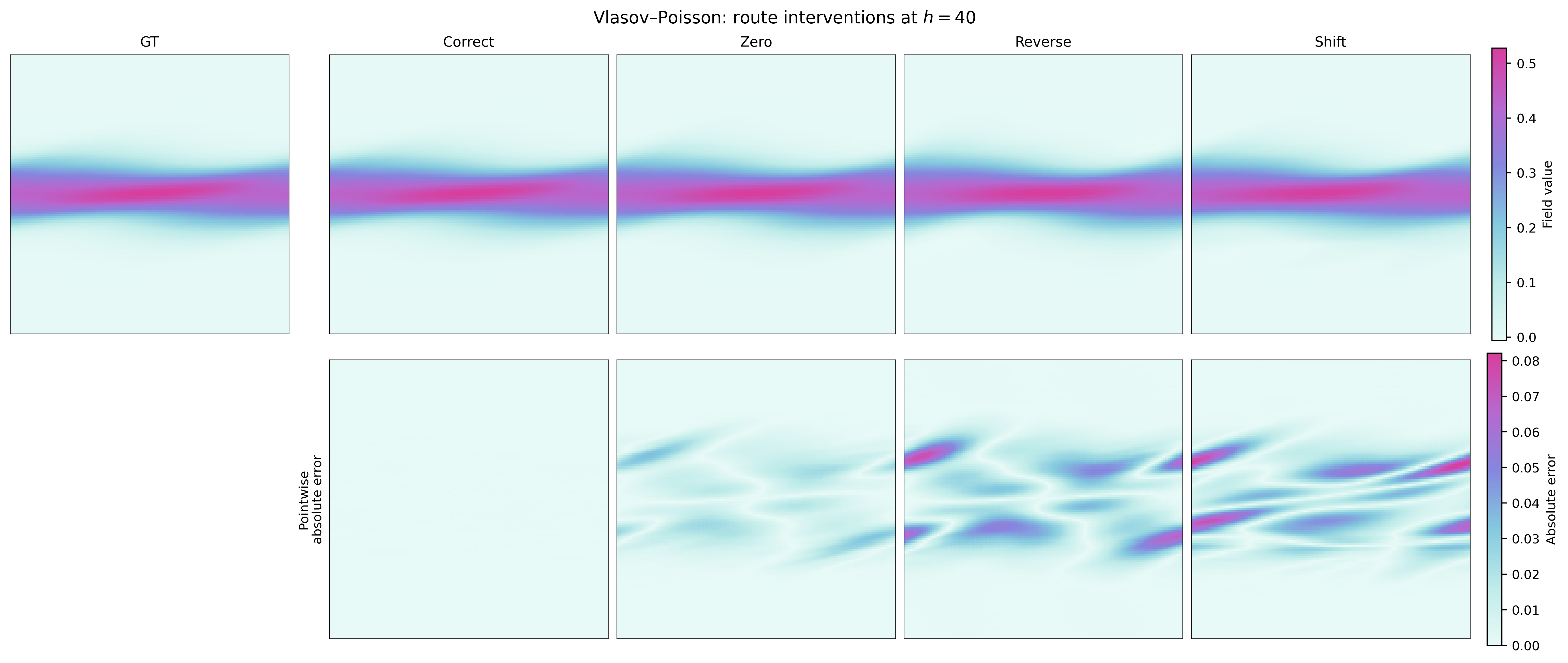}
    \caption{
    \textbf{Route interventions on Vlasov--Poisson at $h=40$.}
    The top row shows the reference phase-space distribution and
    predictions under the \emph{Correct}, \emph{Zero}, \emph{Reverse}, and \emph{Shift} interventions.
    The bottom row shows the corresponding pointwise absolute errors.
    All predictions use the same fitted GRNO checkpoint; only the
    departure route is changed.
    }
    \label{fig:supp_route_vlasov_poisson}
\end{figure}

\begin{figure}[!htbp]
    \centering
    \includegraphics[width=\linewidth]{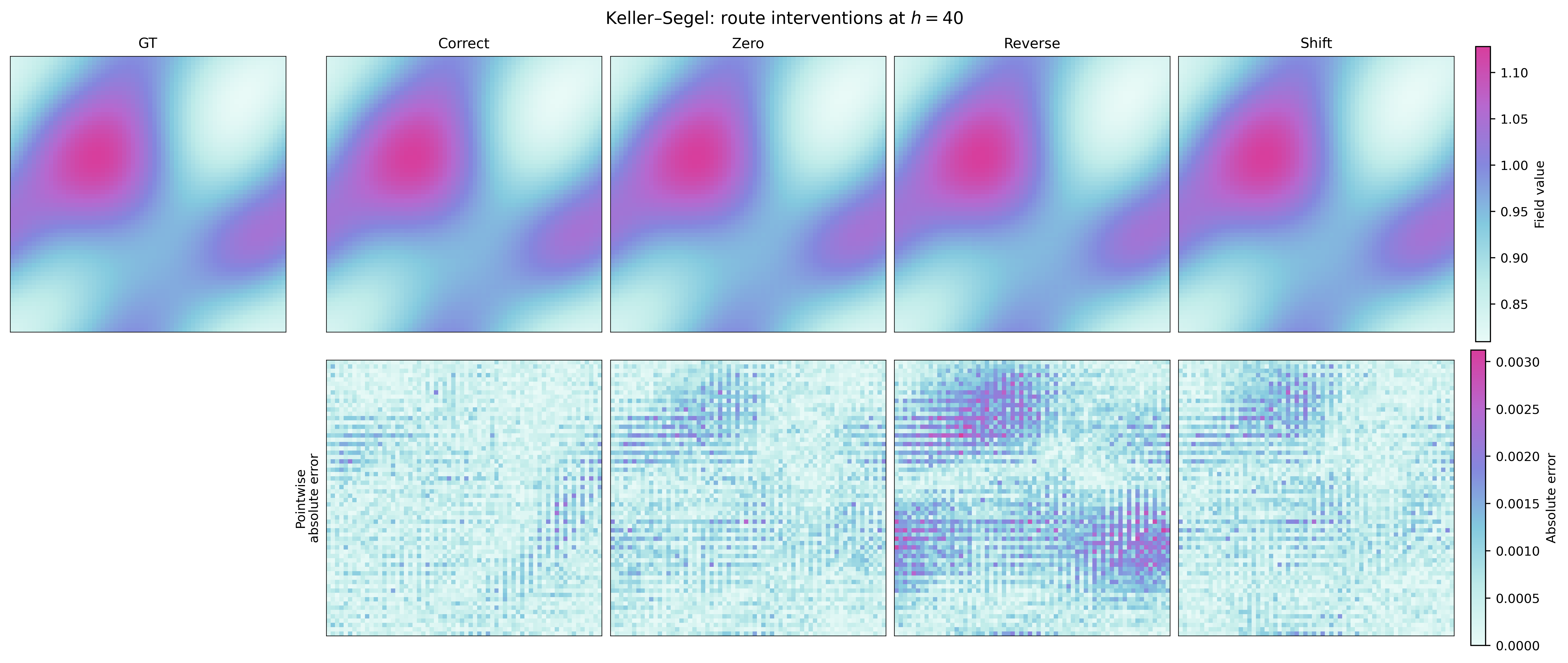}
    \caption{
    \textbf{Route interventions on Keller--Segel at $h=40$.}
    The top row shows the reference field and predictions on the fixed
    visualization slice using the correct, zero, reversed, and shifted
    routes.
    The bottom row shows the corresponding pointwise absolute errors.
    All predictions use the same fitted GRNO checkpoint; only the
    departure route is changed.
    }
    \label{fig:supp_route_keller_segel}
\end{figure}

\begin{figure}[!htbp]
    \centering
    \includegraphics[width=\linewidth]{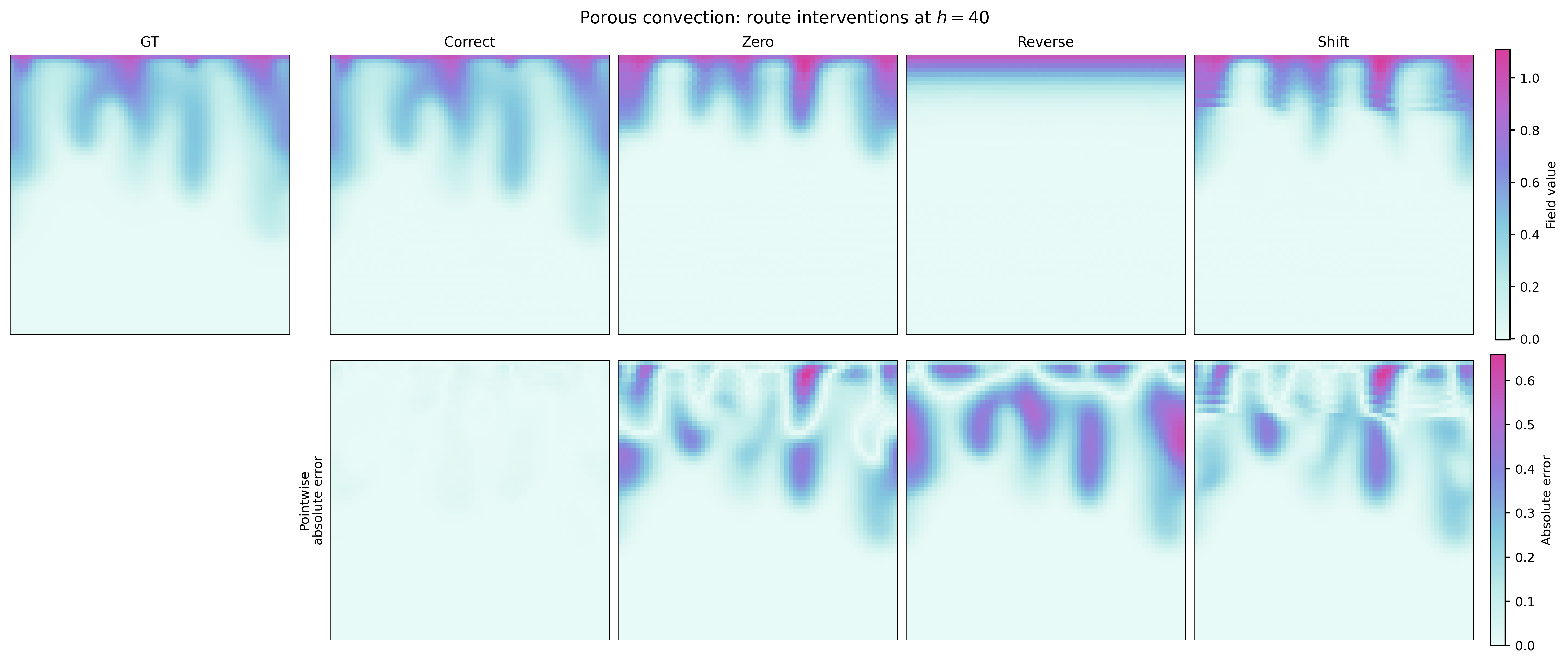}
    \caption{
    \textbf{Route interventions on porous convection at $h=40$.}
    The top row shows the reference concentration field and predictions
    on the fixed visualization slice using the correct, zero, reversed,
    and shifted routes.
    The bottom row shows the corresponding pointwise absolute errors.
    All predictions use the same fitted GRNO checkpoint; only the
    departure route is changed.
    }
    \label{fig:supp_route_porous}
\end{figure}

Figures~\ref{fig:supp_route_kolmogorov}--\ref{fig:supp_route_porous} show the fixed-weight route interventions at $h=40$. For each system, the fitted GRNO checkpoint is held fixed, and only the departure displacement is changed.

\end{document}